\documentclass{article}

\usepackage{microtype}
\usepackage{graphicx}
\usepackage{subcaption}
\usepackage{booktabs} %

\usepackage{hyperref}

\usepackage[preprint]{icml2026}

\usepackage{amsmath}
\usepackage{amssymb}
\usepackage{mathtools}
\usepackage{amsthm}

\usepackage{colortbl}
\usepackage{fancyvrb}
\usepackage[capitalize,noabbrev]{cleveref}

\theoremstyle{plain}
\newtheorem{theorem}{Theorem}[section]

\theoremstyle{definition}
\newtheorem{definition}[theorem]{Definition}

\theoremstyle{remark}

\usepackage{colortbl}

\newcommand{\squishlist}{
\begin{list}{{{\small{$\bullet$}}}}
{\setlength{\itemsep}{3pt}      \setlength{\parsep}{1pt}
\setlength{\topsep}{1pt}       \setlength{\partopsep}{0pt}
\setlength{\leftmargin}{1em} \setlength{\labelwidth}{1em}
\setlength{\labelsep}{0.5em} } }
\newcommand{\squishend}{  \end{list}  }

\newcommand{\method}{{\textbf{MixDPO}}}

\definecolor{darkblue}{rgb}{0, 0, 0.5}
\hypersetup{colorlinks=true, citecolor=darkblue, linkcolor=darkblue, urlcolor=darkblue}

\usepackage[textsize=tiny]{todonotes}

\icmltitlerunning{Small-Margin Preferences Still Matter—If You Train Them Right}

\begin{document}

\twocolumn[
\icmltitle{Small-Margin Preferences Still Matter—If You Train Them Right}

  \icmlsetsymbol{equal}{*}

  \begin{icmlauthorlist}
    \icmlauthor{Jinlong Pang}{ucsc}
    \icmlauthor{Zhaowei Zhu}{docta}
    \icmlauthor{Na Di}{neu}
    \icmlauthor{Yichi Zhang}{rutgers}
    \icmlauthor{Yaxuan Wang}{ucsc}
    \icmlauthor{Chen Qian}{ucsc}
    \icmlauthor{Yang Liu}{ucsc}
  \end{icmlauthorlist}

\icmlaffiliation{ucsc}{University of California, Santa Cruz}
\icmlaffiliation{neu}{Northeastern University}
\icmlaffiliation{docta}{Docta.ai}
\icmlaffiliation{rutgers}{DIMACS, Rutgers University}

\icmlcorrespondingauthor{Yang Liu}{yangliu@ucsc.edu}

  \icmlkeywords{Machine Learning, ICML}

  \vskip 0.3in
]

\printAffiliationsAndNotice{}  %

\begin{abstract}
Preference optimization methods such as DPO align large language models (LLMs) using paired comparisons, but their effectiveness can be highly sensitive to the quality and difficulty of preference pairs. A common heuristic treats small-margin (ambiguous) pairs as noisy and filters them out. In this paper, we revisit this assumption and show that \emph{pair difficulty interacts strongly with the optimization objective}: when trained with preference-based losses, difficult pairs can destabilize training and harm alignment, yet these same pairs still contain useful supervision signals when optimized with supervised fine-tuning (SFT).
Motivated by this observation, we propose \method, a simple yet effective difficulty-aware training strategy that (i) orders preference data from easy to hard (a curriculum over margin-defined difficulty), and (ii) \emph{routes} difficult pairs to an SFT objective while applying a preference loss to easy pairs. This hybrid design provides a practical mechanism to leverage ambiguous pairs without incurring the optimization failures often associated with preference losses on low-margin data.
Across three LLM-judge benchmarks, MixDPO consistently improves alignment over DPO and a range of widely-used variants, with particularly strong gains on AlpacaEval~2 length-controlled (LC) win rate.
\end{abstract}

\section{Introduction}\label{sec:intro}

 Learning from human feedback is essential for aligning large language models (LLMs) with human preferences, helping ensure that these models behave in ways that are helpful, honest, and harmless~\citep{achiam2023gpt, nakano2021webgpt}. A widely adopted approach for such alignment is reinforcement learning from human feedback (RLHF)~\citep{stiennon2020learning, ouyang2022training}, which involves a multi-stage pipeline of LLM fine-tuning and reward model training. 
To simplify this complex process, several off-policy and reward model-free approaches have been proposed, including Direct Preference Optimization (DPO)~\citep{rafailov2024direct} and its variants like KTO~\citep{ethayarajh2024kto} and SimPO~\citep{meng2024simpo}, to name a few.
 These approaches bypass online reinforcement learning by directly training on a fixed dataset of preference pairs $\{(x, y_w, y_l)\}$, where $y_w$ and $y_l$ represent the preferred and dispreferred responses given the prompt $x$, respectively. Unlike the open-ended exploration used in RLHF, these methods rely heavily on the quality of the underlying preference data, which is crucial for achieving strong alignment performance.

Despite considerable efforts having been devoted to curating post-training data, either at the sample level \citep{chen2023alpagasus, xia2024less, pang2024improving, liu2023makes} or the token level \citep{lin2024rho, pang2025token}, the role of data in preference alignment remains heavily overlooked and underexplored.
Building on the idea of data selection, recent work \citep{deng2025less, huang2025larger, gao2025principled} investigates pairwise data quality based on the margin between preferred and dispreferred responses within each pair, aiming to identify easy, large-margin pairs. In contrast, those \textit{difficult} (small-margin) pairs are often discarded based on the intuition that they introduce ambiguity or noise, hindering effective preference modeling.

This rationale stems from the observation that such difficult pairs often induce a \emph{likelihood displacement} phenomenon \citep{pal2024smaug, yuan2024advancing, rafailov2024r, razin2024unintentional}, where the log probability of both the preferred response $y_w$ and dispreferred response $y_l$ decreases, contradicting the initial goal of preference learning.
The potential reason is that subtle differences between similar responses can lead to unstable gradients and misaligned optimization. Most existing approaches attempt to alleviate this issue by redesigning DPO-style loss functions, such as Cal-DPO \citep{xiao2024cal} and DPOP \citep{pal2024smaug}. In contrast, from a data-centric perspective, we raise an interesting question that \textit{“Can these difficult pairs, which are typically discarded, still offer valuable supervision signals if properly utilized?”}

\begin{figure}[t]
    \centering
        \includegraphics[width=1.02\linewidth]{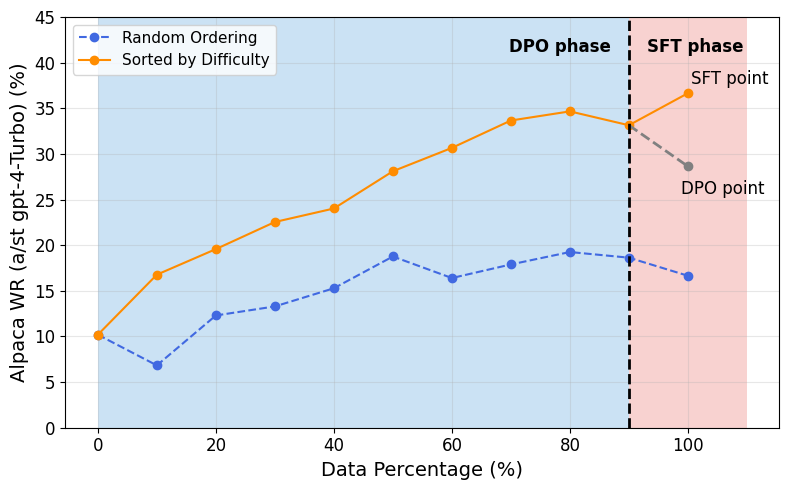}
\caption{Performance comparison between random ordering and difficulty-based sorting on the AlpacaEval 2 benchmark. DPO is used as the default loss function. Notably, instead of discarding difficult preference pairs, further training them using the SFT loss leads to improved performance.}
    \label{fig:sft_benefit}
\end{figure}

In this work, we challenge the conventional practice by proposing a new insight that 
\textit{“Despite difficult pairs may hinder alignment when optimized with preference-based objectives due to potential likelihood displacement, they can still provide valuable learning signals when trained with supervised fine-tuning (SFT).”}

Our empirical experiments in Figure~\ref{fig:sft_benefit} support two key findings:
(1) Given the natural variation in difficulty across preference pairs, structuring training from easy to difficult pairs, compared with random ordering, leads to better alignment performance, consistent with the curriculum learning paradigm;
(2) While difficult pairs often harm preference-based optimization, they become beneficial when trained using SFT loss.
Ultimately, backed by empirical evidence, we introduce a simple yet effective method that adaptively switches from DPO to standard SFT loss based on the difficulty of preference pairs.
Specifically, we apply DPO loss to optimize easier pairs, while resorting to SFT loss for the more challenging ones to better harness their informative value. Importantly, SFT loss provides a straightforward way to avoid likelihood displacement.

We summarize our main contributions as follows.
\squishlist
    \item We conduct a systematic study on the role of easy and difficult preference pairs in alignment performance, revealing a consistent pattern: performance deteriorates as training data shifts from easy to difficult pairs.

    \item Challenging the common practice of discarding difficult pairs with small score margins, we propose a novel insight—these pairs can still offer valuable learning signals when optimized using supervised fine-tuning (SFT), as confirmed by empirical evidence. Motivated by this, we introduce a simple yet effective method, \method, which applies DPO loss to easy pairs and SFT loss to difficult ones to better utilize informative signals.

    \item We perform extensive experiments to validate the effectiveness of \method, against DPO and several widely-used variants, including CPO, IPO, KTO, SimPO, and SelectiveDPO. Notably, the introduced SFT phrase serves to counteract the tendency toward overly long responses driven by preference bias, as evidenced by improvements in the AlpacaEval 2 LC win rate.
     The extensive ablation study on the additional base model and Agrilla-7k dataset demonstrates the generality and adaptability of our approach.

\squishend

\section{Related Work}\label{sec:related_work}

\paragraph{LLM Preference Alignment} Start from work \citep{ouyang2022training}, numerous approaches have been proposed to align LLM-generated responses with human preferences. These methods can be broadly categorized into two paradigms: RLHF and DPO \citep{rafailov2024direct}. As a simplified, reward model-free alternative to RLHF, DPO has emerged as one of the most widely adopted techniques for preference alignment. A growing body of research has focused on analyzing the DPO loss from various theoretical and practical perspectives, leading to several notable variants, including IPO \citep{azar2024general}, KTO \citep{ethayarajh2024kto}, CPO \citep{xu2024contrastive}, and SimPO \citep{meng2024simpo}, to name a few.

While substantial effort has been devoted to data selection in instruction tuning and pre-training at both the sample level \citep{chen2023alpagasus, xia2024less, zhang2025evaluating, pang2024improving, yang2025entp, liu2023makes, liu2024automatic} and the token level \citep{lin2024rho, pang2025token}, the data impact on the preference alignment remains overlooked and underexplored.
Several prior studies \citep{bai2022training, pang2024fairness, ethayarajh2110understanding} have examined the influence of data during alignment, but primarily from the perspective of aligning models with human values and ethics. Several recent work \citep{deng2025less, huang2025larger, gao2025principled} investigates data quality based on different metrics such as reward score, aiming to retain informative samples while discarding those deemed uninformative or noisy. Another line of work focuses on developing noise-tolerant DPO objectives, such as cDPO \citep{mitchellnote}, robustDPO \citep{chowdhury2024provably}, and PerpCorrect \citep{kongperplexity}, but does not investigate the impact of individual samples.
In contrast, our work challenges this binary view of data quality. We argue that seemingly uninformative or overly difficult samples that are typically filtered out in preference-based optimization, can still provide valuable information when optimized using SFT objectives.

\paragraph{Curriculum Learning}
Curriculum learning (CL) follows the natural human learning patterns by structuring learning from simpler to more complex concepts \citep{avrahami1997teaching, bengio2009curriculum}, which could effectively accelerate model convergence and enhance generalization. Inspired by its success, CL patterns have been incorporated into several domains, including machine translation \citep{platanios2019competence}, image generation \citep{croitoru2024reverse}, and multimedia search \citep{jiang2014easy, tudor2016hard}.

In preference alignment for LLMs, a central component of curriculum learning is the definition of sample difficulty on preference pairs. Recent studies have explored various difficulty scoring metrics to enable curriculum-based training, such as prompt length or inherent attention scores \citep{kim2024strategic}, model perplexity on responses \citep{wu2024curriculum}, reward margins estimated by strong reward models \citep{croitoru2024curriculum}, and validation DPO loss on preference pairs \citep{gao2025principled}. In this work, we adopt a simple yet effective strategy by using the original rating scores available in the preference data to estimate difficulty, incurring no additional computational overhead. Moreover, while \citet{gao2025principled} opts to filter out overly difficult examples due to their potential negative impacts, we present an alternative approach, demonstrating that these samples can still contribute to alignment when optimized using their SFT loss.

\section{Preliminary}\label{sec:preliminary}

\subsection{Direct Preference Optimization (DPO)}

Preference alignment \citep{ouyang2022training} aims to ensure that LLMs generate outputs that reflect human preferences and communication styles, thereby enhancing their safety, reliability, and trustworthiness in real-world settings.
In this work, we adopt DPO \citep{rafailov2024direct}, one of the most widely used methods for preference-based alignment. Given a pairwise preference dataset $\mathcal{D} := {(x, y_w, y_l)}$, where $x$ is a prompt, $y_w$ is the preferred response, and $y_l$ is the less preferred response, DPO trains a policy model $\pi_{\theta}$ using the following objective:
\begin{align*}
    & \mathcal{L}_{\text{DPO}}(\pi_\theta; \pi_{\text{ref}}) = \\ & \qquad  -\mathbb{E}_{(x, y_w, y_l) \sim \mathcal{D}} \left[ \log \sigma\left( r_{\theta}(x,y_w) - r_{\theta}(x,y_l) \right) \right], 
\end{align*}

where $r_{\theta}(x,y) = \beta \log \frac{\pi_\theta(y \mid x)}{\pi_{\text{ref}}(y \mid x)}$, $\pi_{\text{ref}}$ represents the reference policy model—typically the model fine-tuned via SFT, $\sigma(\cdot)$ denotes the sigmoid function, and $\beta$ is a hyperparameter that controls the divergence between the reference model $\pi_{\text{ref}}$ and the current policy $\pi_{\theta}$.

\begin{figure*}[t]
    \centering
    \includegraphics[width=0.91\linewidth]{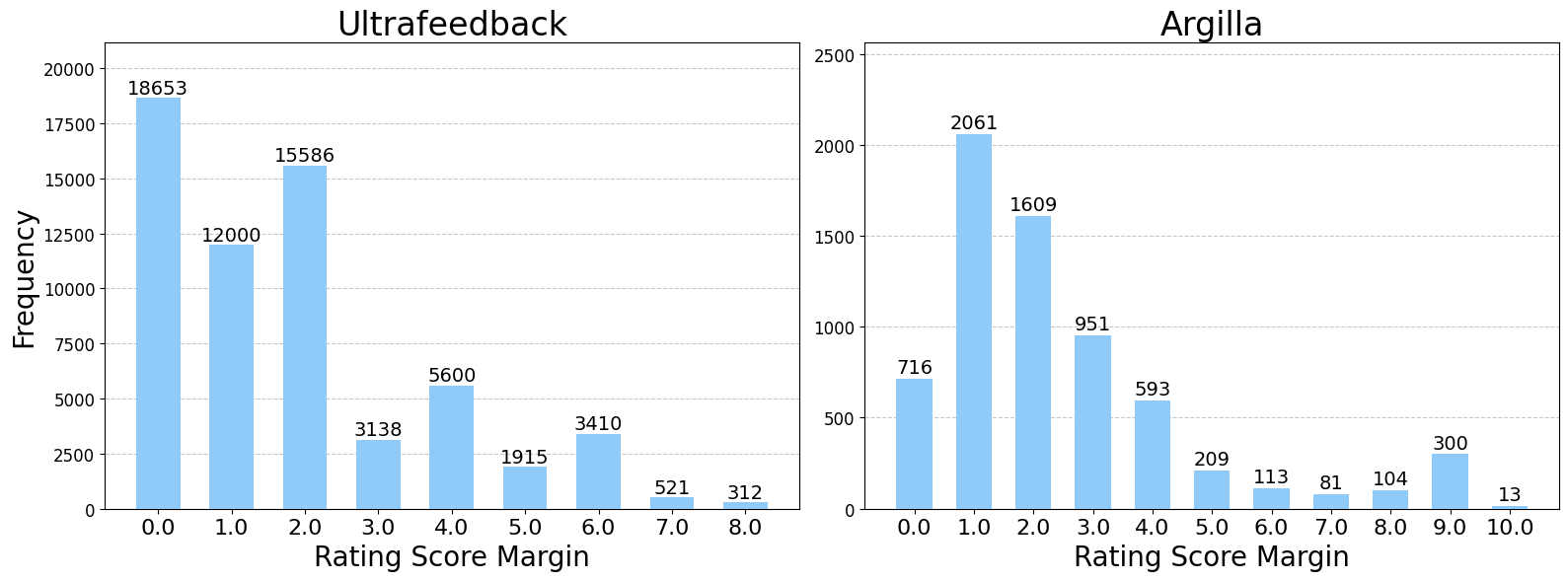}
    \caption{Original rating score margin distribution. \textbf{Left}: Ultrafeedback dataset (61k samples). \textbf{Right}: Argilla dataset (7k samples). Chosen/Rejected scores are both annotated from various LLMs.  Observe that approximately 50\% of the samples in both datasets exhibit a score difference below 1.0.}
    \label{fig:score_difference_distribution}
\end{figure*}

\subsection{Curriculum Learning}

Curriculum learning \citep{bengio2009curriculum} is a training strategy that gradually exposes the model to increasingly difficult examples, inspired by how humans learn. Curriculum learning can improve convergence speed and model generalization by starting with easier samples and progressively moving to harder ones.
This paradigm relies on assessing sample difficulty to guide the model from learning on easier examples to more challenging ones over time.

\paragraph{Qualifying pairwise difficulty} 
Note that preference datasets used for DPO are typically constructed from response pairs annotated with LLM-generated rating scores. In practice, for each prompt, the response with the higher rating score is selected as the chosen response, while the one with the lower score is designated as rejected.
Building upon this construction, the rating scores can naturally serve as a basis for estimating the sample-level difficulty. We define this notion as follows:

\begin{definition}\label{def:difficulty}
\textbf{(Pairwise Difficulty)}
Given a preference pair represented as $(x, y_w, y_l, s_w, s_l)$, where $s_w$ and $s_l$ denote the rating scores of the chosen and rejected responses, respectively, the pairwise difficulty is defined as the rating score difference:
$M(s_w, s_l) := s_w - s_l$.

\end{definition}
In practice, metric $M(\cdot)$ is always non-negative.
By default, the rating score margin ($M$) will be utilized to measure pairwise difficulty, where a higher $M$ implies an easier pair.
Intuitively, preference pairs within a dataset can differ in difficulty, as reflected by the varying rating score margins between the chosen and rejected responses.
Figure~\ref{fig:score_difference_distribution} presents the score gap distributions of two popular preference datasets, including Ultrafeedback and Arigilla, highlighting the widespread presence of difficult pairs in the dataset.

While Definition \ref{def:difficulty} serves as our primary metric, alternative metrics can also be considered, such as reward score margin \citep{croitoru2024curriculum}, DPO loss \citep{gao2025principled}, or the embedding distance between responses. Note that our goal is not to develop a superior metric for difficulty estimation. Instead, we investigate the role of difficulty based on the available annotations used during dataset construction. Notably, rating scores are computation-free, whereas model-dependent measures like DPO loss are more costly. More discussion can be found in Section~\ref{sec:related_work}. For difficult pairs, we provide several examples extracted from the Ultrafeedback dataset \citep{cui2023ultrafeedback} in Appendix \ref{apx:difficult_pairs_examples}.

\section{Exploring the Impact of Easy/Difficult Pairs}\label{sec:method}

In this section, we begin by empirically validating the potential impact of easy and difficult preference pairs on alignment performance. Given the observed results, we further explore strategies for handling these difficult pairs.  
While existing approaches often opt to filter out such pairs to mitigate their negative effects \citep{gao2025principled, deng2025less}, we argue that, when properly utilized, these difficult pairs remain informative and can contribute positively to alignment performance through our proposed method.  Instead of discarding such pairs directly, we introduce a hybrid objective that applies DPO loss to easy pairs and employs SFT loss for the more challenging ones.

\subsection{Empirical Evidence: Potential Impact of Easy or Difficult Pairs}

\begin{figure*}[h]
    \centering
    \hspace{-0.1in}
        \begin{minipage}[b]{0.25\linewidth}
        \centering
    \includegraphics[width=\linewidth]{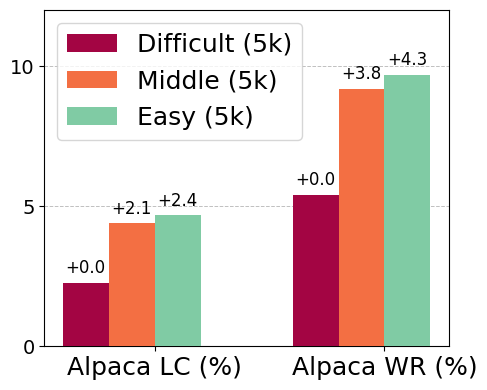}
    \subcaption{Alpaca LC-WR}
    \label{fig:win_rate_compare}
    \end{minipage}%
    \begin{minipage}[b]{0.25\linewidth}
        \centering
    \includegraphics[width=\linewidth]{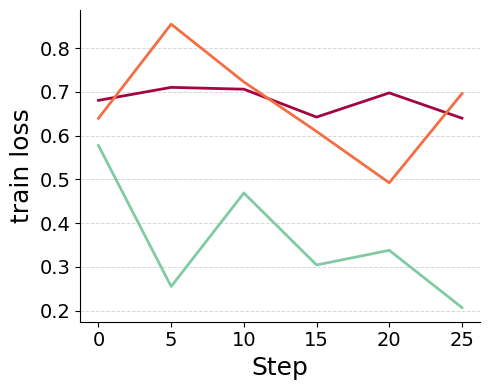}
        \subcaption{Train loss}
        \label{fig:train_loss_compare}
    \end{minipage}%
    \begin{minipage}[b]{0.25\linewidth}
        \centering
        \includegraphics[width=\linewidth]{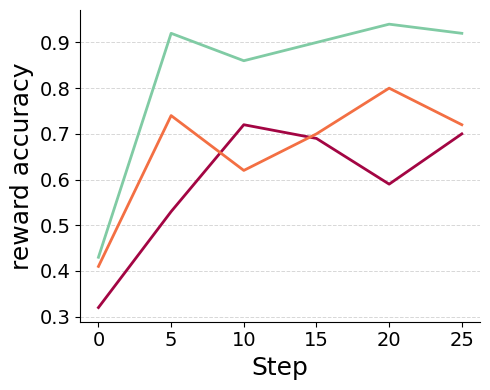}
        \subcaption{Reward accuracy}
        \label{fig:reward_accuracy_compare}
    \end{minipage}%
    \begin{minipage}[b]{0.25\linewidth}
        \centering
        \includegraphics[width=\linewidth]{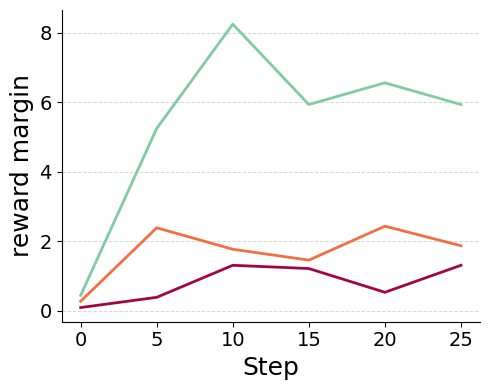}
        \subcaption{Reward margin}
        \label{fig:reward_margin_compare}
    \end{minipage}
    \caption{Performance comparison of models fine-tuned on three subsets of the UltraFeedback dataset, each representing a different difficulty level. The base model used is LLaMA-3-8B, and DPO is adopted as the default loss function. Subfigure~\ref{fig:win_rate_compare} highlights the positive influence of easy pairs on alignment performance. Subfigures~\ref{fig:train_loss_compare}-\ref{fig:reward_margin_compare} provide detailed training dynamics, revealing a clear trend: as the data shifts from easy to difficult pairs, the performance consistently declines.}
    \label{fig:easy_difficult_comparison}
\end{figure*}

To investigate the impact of data pairs with varying difficulty levels, we partition the UltraFeedback dataset into three subsets based on the distribution of rating score differences assigned by the LLM. For each subset, we randomly sample 5,000 examples and apply the typical DPO loss for training.
Figure~\ref{fig:easy_difficult_comparison} illustrates how alignment performance varies across these levels. Remarkably, the performance results shown in Figure~\ref{fig:win_rate_compare} illustrate that easier preference pairs—those with larger score gaps—lead to higher scores in both AlpacaEval 2's LC win rate and raw win rate, whereas performance degrades when training on more difficult pairs with smaller score differences. 
In support of this trend, Figures~\ref{fig:train_loss_compare}-\ref{fig:reward_margin_compare} provide more detailed training process information, showing that easy pairs result in faster convergence (lower train loss), higher reward accuracy, and larger reward margins. In contrast, difficult pairs lead to slower optimization, weaker preference signals, and limited reward separation, ultimately reducing alignment performance.

Motivated by the above observation, we sort preference pairs by difficulty and prioritize those with larger score margins (i.e., easier pairs), following a curriculum-style training strategy \citep{bengio2009curriculum, kim2024strategic, wu2024curriculum}. 
The rationale behind this strategy is that easier pairs provide clearer and more consistent supervision signals, which can effectively amplify the divergence between the base and reference models and further enhance DPO optimization. However, how to further handle these remaining difficult pairs raises a critical question, given the observed negative impact on alignment performance \citep{gao2025principled, deng2025less}. 

\paragraph{Filtering out or not?} A common practice in prior work is to discard difficult pairs entirely \citep{deng2025less, gao2025principled}, under the intuition that they introduce ambiguity or extra noise that can hinder effective preference modeling. 
This concern is rooted in the observation that such hard-to-distinguish or similar pairs often induce a \emph{likelihood displacement} phenomenon \citep{pal2024smaug, yuan2024advancing, rafailov2024r, razin2024unintentional}, where the log probability of both the preferred response $y_w$ and dispreferred response $y_l$ decreases, which contradicts the initial design of DPO that increase the probability of $y_w$ and decrease the probability of $y_l$.
The potential reason is that small differences between two nearly identical responses can lead to unstable gradient signals and misaligned optimization. More details can be found in Appendix~\ref{apx:compare_with_likelihood_displacement_method}.

Instead of discarding those difficult pairs, in this work, we argue that these difficult pairs are still informative; they can provide valuable supervision signals if handled properly. For example, recent work \citep{xu2024dpo} illustrates that filtering out difficult pairs can be detrimental to overall DPO performance across LLM truthworthy tasks, aligning with our insight. In the following section, we delve into how to handle those difficult pairs, if retained.

\subsection{Objective: Apply DPO to Easy Pairs, SFT to Difficult Ones}
Inspired by the recent work \citep{grattafiori2024llama, pal2024smaug}, which suggests adding SFT-like terms to mitigate likelihood displacement, we propose a simple yet effective approach that adaptively switches between DPO and SFT loss depending on pair difficulty.

Concretely, we apply DPO loss to easy pairs with large score margins, where model preferences are more confident, and use SFT loss on difficult pairs with small margins to avoid optimization instability. This margin-aware objective combines the alignment strength of DPO with the robustness of SFT.
These design choices form the basis of our proposed method, \method, which we describe in detail below. We define a binary indicator $z$ to distinguish easy from difficult pairs and conditionally apply the corresponding loss as follows:
\begin{equation*}
    z = 
    \begin{cases}
        1 & \text{if } M(s_w, s_l) < \tau \\
        0 & \text{otherwise}
    \end{cases}
\end{equation*}
where $\tau$ is a predefined threshold controlling the difficulty sensitivity. By default, we set $\tau$ as 0.5 for the Ultrafeedback dataset. Based on the indicator $z$, we construct a hybrid loss that applies DPO to easier pairs ($z = 0$) and SFT to difficult ones ($z = 1$):
\begin{align*}
        \mathcal{L}_{\method} = 
    - & (1 - z) \cdot \underbrace{\log p_{\theta} (y_w \succ y_l \mid x)}_{\text{DPO loss}}  \\
    & - z \cdot \underbrace{\frac{1}{T}\sum_{t=1}^{T} \log p_\theta(y_{w, t} \mid x, y_{w, <t})}_{\text{SFT loss}}
\end{align*}

Here, $z$ serves as a switching signal that dynamically assigns the appropriate loss to each training example based on its difficulty.

Note that preference pairs are sorted according to their estimated difficulty. In practice, we adopt a two-stage training paradigm: standard DPO training is first applied to the easier pairs with larger score margins, followed by an SFT fine-tuning phase that focuses on the more difficult pairs.

\section{Experiments}\label{sec:exp}

\subsection{Experimental Setup}

\paragraph{Base models \& training settings} Following the setting of SimPO \citep{meng2024simpo}, we utilize two popular models as our base models: LLaMA-3-8B \citep{grattafiori2024llama} and Mistral-7B-v0.1 \citep{jiang2023mistral}. In particular, the finetuned version of these two models on the Ultrachat-200k dataset\footnote{\url{https://huggingface.co/datasets/HuggingFaceH4/ultrachat_200k}} is used as the starting point for the following preference optimization.
We perform preference optimization on the Ultrafeedback dataset \citep{cuiultrafeedback} for evaluation.
For the UltraFeedback dataset, we set $\tau=0.5$, resulting in 7,387 identified difficult pairs, with the remaining pairs classified as easy. In Appendix~\ref{sec:more_experiment_results}, additional results are provided to further examine the impact of the threshold $\tau$.

\begin{table*}[ht]
    \caption{Performance comparison of different baselines on three LLM judge benchmarks. Note that $\widetilde{\textbf{Win}}$ represents the adjusted win rate, which equals the win rate plus half of the tie rate. We highlight the best results in \textbf{boldface} and the second-best with \underline{underline}.  }
    \centering
    \resizebox{0.9\linewidth}{!}{
    \begin{tabular}{c|cc|cc|c}
    \toprule
    ~ & \multicolumn{2}{c|}{\textbf{AlpacaEval 2.0}} & \multicolumn{2}{c|}{\textbf{Arena Hard}}  & \textbf{MT\_Bench} \\
    \cmidrule(lr){2-3} \cmidrule(lr){4-5} \cmidrule(lr){6-6}
    \textbf{Method}  & \textbf{LC Win Rate (\%)} & \textbf{Win Rate (\%)} & \textbf{Win  Rate (\%)} & $\widetilde{\textbf{Win}}$ \textbf{Rate (\%)} & \textbf{Avg. Score (0-10)} \\
    \midrule
    \multicolumn{6}{c}{\cellcolor{blue!10} \textbf{Base model: LLaMA-3-8B} } \\
    \midrule
    \textsc{LLaMA3-8B-Base-SFT} & 3.73 & 10.19 & 3.9 & 7.8 & 4.82 \\
    \textsc{Vanilla DPO} & 9.37 & 16.77 & 20.4 & 31.2 & 5.94 \\
    \midrule
    \textsc{SLiC-HF} & 5.20  &  5.71 & 14.4 & 22.8 & 4.99 \\
    \textsc{CPO} & 4.25 & 9.69 & 12.8 &24.0 & 5.64 \\
    \textsc{IPO} & 5.89 & 11.55 & \textbf{20.6}& \textbf{32.5}& 6.01 \\
    \textsc{KTO} & 4.27 & 3.98 & 17.6 &27.4 & 6.03 \\
    \textsc{O-RPO} & 5.43   &   7.08 & 14.4 &24.0 & 5.80 \\
    \textsc{RDPO} & 6.92  &   11.06 &  18.9 &28.6 & 5.97 \\
    \textsc{SimPO} & 6.77 & 14.04 & 20.2 &\textbf{32.5} & \underline{6.09} \\
    \textsc{SelectiveDPO} & \underline{8.85} & \underline{30.43} & \underline{20.5} &\underline{32.1} & 5.82 \\
    \midrule
    \method & \textbf{14.42} & \textbf{36.65} & 16.6 & 26.3 & \textbf{6.17} \\ %

    \toprule
    \multicolumn{6}{c}{\cellcolor{blue!10} \textbf{Base model: Mistral-7B-v0.1}} \\
    \midrule
    \textsc{Mistral-7B-Base-SFT} & 2.39 & 1.24 & 3.0 &5.1 & 4.53 \\
    \textsc{Vanilla DPO} & 5.14 & 4.72 & 10.0& 15.0 & 5.14 \\
    \midrule
    \textsc{SLiC-HF} & 4.42   &   3.60  & 6.0& 10.7 & 4.43\\
    \textsc{CPO} & 4.04 & 3.85 & 4.6 &10.2 & 4.6 \\
    \textsc{IPO} & 5.45 & 4.60 & 6.8 &13.2 & 4.73 \\
    \textsc{KTO} & 5.02 & 3.23 & 5.0& 10.3 & 4.56 \\
    \textsc{O-RPO} &   4.38 &     3.35   & 2.8 &4.6 & 2.74 \\
    \textsc{RDPO} &   \underline{6.03}    &  4.60    &  9.7& 16.7& 5.29 \\
    \textsc{SimPO} & 4.30 & \underline{5.47} & \textbf{11.2}& \underline{19.5} & \underline{5.34} \\
    \textsc{SelectiveDPO} & 3.91    &  \underline{5.47}  & \underline{10.2} &16.6 & 4.98 \\ %

    \midrule
    \method   & \textbf{7.67}    &  \textbf{6.71} & \underline{10.2} & \textbf{20.5}  &  \textbf{5.55} \\ %

    \bottomrule
    \end{tabular}
    }
    \label{tab:main_results}
\end{table*}

\paragraph{Baselines}
We evaluate the performance of the proposed preference method by benchmarking state-of-the-art preference optimization methods, including DPO \citep{rafailov2024direct} and its variants such as SLiC-HF \citep{zhao2023slic}, IPO \citep{azar2024general}, KTO \citep{ethayarajh2024kto}, CPO \citep{xu2024contrastive}, SimPO \citep{meng2024simpo}, O-RPO \citep{hong2024orpo}, R-DPO \citep{park2024disentangling}, and SelectiveDPO \citep{gao2025principled}.  For consistency, we utilize the released models from the SimPO repository\footnote{\url{https://github.com/princeton-nlp/SimPO}} to generate model responses and then conduct evaluation. For SelectiveDPO, we follow the original setup and use the released validation DPO loss as its difficulty metric.
Following the same training setup, by default, all results are based on full parameter fine-tuning (FPFT).  More hyperparameter settings could be found in Appendix \ref{sec:apx_experiment_details}.

\paragraph{Evaluation benchmarks} In this paper, we primarily select three popular open-ended benchmarks: MT-Bench \citep{zheng2023judging},  AlpacaEval 2 \citep{li2023alpacaeval}, and Arena-Hard-v0.1 \citep{li2024live}. These benchmarks evaluate the models' versatile conversational abilities across diverse queries and have been widely adopted by the community. For example, AlpacaEval 2 consists of 805 questions from 5 subsets (e.g., SelfInstruct, Vicuna-Bench). Following their evaluation protocols, we report the Length-Control win rate (LC Win Rate) and win rate for the AlpacaEval 2 against GPT-4-Turbo, win rate for Arena-Hard against GPT-4-0314, and a discrete score (0-10) for the MT-Bench benchmark. Owing to cost considerations, we adopt the recently released GPT-4.1 (i.e., GPT-4.1-2025-04-14) as our LLM judge model across all benchmarks.

\subsection{Empirical Results}

As shown in Table~\ref {tab:main_results}, our proposed method consistently outperforms baselines across two base models (i.e., LLaMA-3-8B and Mistral-7B-v0.1) on three evaluation benchmarks.
Remarkably, under the LLaMA-3-8B setting, our proposed method achieves an absolute performance gain of approximately 6\% in both the LC win rate and the raw win rate compared with all baselines. 
Similar performance improvements have also been observed under the Mistral-7B-v0.1 setting. These consistent improvements underscore the effectiveness of our proposed approach. Across both base model settings, one can observe that many DPO variants, such as CPO and KTO, fail to outperform, and in some cases underperform standard DPO, consistent with observations reported in SimPO \citep{meng2024simpo}, which used more powerful GPT-4 for LLM judgement.

\section{Ablation Study}

In this section, we conduct a comprehensive ablation study to evaluate the effectiveness and generalizability of our proposed method on additional base models and preference datasets. We also deeply analyze the contribution of each component, assess its compatibility with existing DPO variants, and report performance results on the stronger LLaMA-3-8B-Instruct base model.

\begin{table*}[ht]
\centering
\caption{Performance comparison across different settings. \textbf{Left}: Evaluating generalization to the \underline{Qwen-2.5-7B} base model using the UltraFeedback dataset. \textbf{Right}: Evaluating generalization to a different preference dataset, Argilla-7k, with the LLaMA-3-8B model.}
\begin{minipage}[t]{0.47\linewidth}
    \centering
    \resizebox{\linewidth}{!}{
    \begin{tabular}{ccc}
    \toprule
    ~ & \multicolumn{2}{c}{\cellcolor{blue!10} \textbf{Model: Qwen2.5-7B}}\\
    \cmidrule(lr){2-3}
    \textbf{Method} & \textbf{LC Win Rate (\%)} & \textbf{Win Rate (\%)} \\
    \midrule
    \textsc{Base SFT}  & 0.20   &   0.99 \\
    \midrule
    \textsc{vanilla DPO} & 2.38   &   3.60 \\
    \textsc{SimPO} &     2.28      &     3.11   \\
    \textsc{SelectiveDPO} & 3.12  &    \textbf{5.59} \\
    \midrule
    \method & \textbf{3.45}   &   \textbf{5.59} \\
    \bottomrule
    \end{tabular}}
    \label{tab:qwen_results}
\end{minipage}%
\hfill
\begin{minipage}[t]{0.47\linewidth}
    \centering
    \resizebox{\linewidth}{!}{
    \begin{tabular}{ccc}
    \toprule
    ~ & \multicolumn{2}{c}{\cellcolor{blue!10} \textbf{Dataset: Argilla-7k}} \\
    \cmidrule(lr){2-3}
    \textbf{Method} & \textbf{LC Win Rate (\%)} & \textbf{Win Rate (\%)} \\
    \midrule
    \textsc{Base SFT}  & 3.73    &  10.19 \\
    \midrule
    \textsc{vanilla DPO} & 2.90   &  12.17\\
    \textsc{SimPO} & 7.07     &  5.84 \\
    \textsc{SelectiveDPO} & 3.59      &         5.22 \\
    \midrule
    \method & \textbf{9.23 } &   \textbf{20.62} \\
    \bottomrule
    \end{tabular}}
    \label{tab:argilla_results}
\end{minipage}
\end{table*}

\subsection{Generalization Across Models and Datasets}

Here, we investigate the generalization of our proposed method, \method, on an additional base model and preference dataset. From the baseline pool, we selectively choose two strong-performing DPO variants: SimPO and SelectiveDPO. Given their potentially complex hyperparameter settings and sensitivity (e.g., to learning rate), we make our best effort to tune them for optimal performance. More details on the hyper-parameter configurations are provided in Appendix~\ref{sec:more_experiment_results}. Here, we use Qwen-2.5-7B as the base model and Argilla-7K dataset\footnote{\url{https://huggingface.co/datasets/argilla/dpo-mix-7k}} to evaluate the generalization of our method.

\vspace{-1ex}
\paragraph{\method~performs well on Qwen-2.5-7B model} To validate the effectiveness of our proposed method on another model suite, we select Qwen-2.5-7B \citep{yang2024qwen2} as our third base model. Table~\ref{tab:qwen_results} (Left) demonstrates the corresponding results on the AlpacaEval 2 benchmark. Observe that our proposed method outperforms baselines in both LC win rate and raw win rate.

\vspace{-1ex}
\paragraph{\method~generalizes well to the Argilla-7K dataset}
We evaluate the generalization capability of our method on a different dataset, Argilla, which comprises 7k high-quality preference samples aggregated from multiple sources. The results in Table~\ref{tab:argilla_results} (Right), demonstrate that \method~consistently outperforms all baselines, underscoring its robustness and generalization across datasets.

\subsection{Understanding and Extending MixDPO: Component and Variant Analysis}

\begin{figure*}[ht]
    \centering
    \begin{minipage}[t]{0.47\linewidth}
        \centering
        \includegraphics[width=\linewidth]{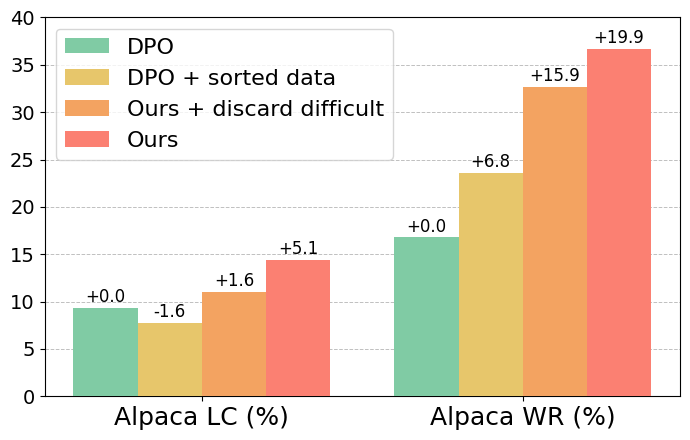}
        \subcaption{Contribution of each component used in MixDPO.}
        \label{fig:seperate_impact_of_component}
    \end{minipage}
    \hfill
    \begin{minipage}[t]{0.47\linewidth}
        \centering
        \includegraphics[width=\linewidth]{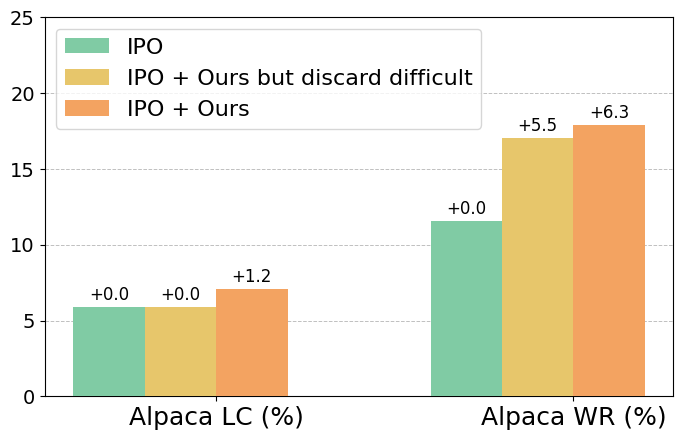}
        \subcaption{Integrating MixDPO into IPO.}
        \label{fig:combine_with_dpo_variants}
    \end{minipage}
    \caption{Ablation results based on the LLaMA-3-8B model. Left: Contribution of each component within MixDPO. Right: Effect of integrating MixDPO into existing DPO variant IPO. }
    \label{fig:ablation_combined}
\end{figure*}

\paragraph{Exploring the separate efforts of different components}
Note that our proposed method comprises two key components: sorting data by difficulty and applying a hybrid loss function. To highlight the contributions of each component explicitly, Figure~\ref{fig:seperate_impact_of_component} presents the performance of various configurations based on the LLaMA-3-8B model. 
Notably, the DPO training stage in the last three settings—\texttt{DPO+sorted data} and \texttt{Ours+discard difficult}—is identical, using the same loss and the same difficulty-sorted data. They differ only in the final SFT stage: \texttt{DPO+sorted data} continues training difficult pairs with DPO, \texttt{Ours+discard difficult} drops them and stops training, and \texttt{Ours} further trains on them using the SFT objective.
The empirical results show that both components independently improve alignment performance. Importantly, rather than discarding difficult pairs, leveraging them through SFT loss leads to a notable improvement in the AlpacaEval 2 LC win rate, without substantially compromising the raw win rate.

\paragraph{Can \method~be applied to other DPO variants?}
We are also interested in investigating whether MixDPO can be incorporated into other DPO variants. Here, we selectively examine one DPO variant, IPO.
Following our framework, we first sort preference pairs based on the score margin. For IPO, we substitute the DPO loss with the IPO loss when training on easy pairs, while applying the SFT loss to the difficult ones.  For consistency, we select the last 7,387 difficult pairs ranked by pairwise difficulty and train them using SFT loss, following our proposed method.
Figure~\ref{fig:combine_with_dpo_variants} compares the performance of each variant before and after applying our approach. Notably, integrating our method consistently improves the original alignment performance, demonstrating its broad applicability and effectiveness. More empirical results can be found in Appendix~\ref{sec:more_experiment_results}.

\paragraph{Performance on stronger instruction-tuned model.}
To further demonstrate the effectiveness of MixDPO on stronger instruction-tuned models, we follow the experimental setups of SimPO \citep{meng2024simpo} and SelectiveDPO \citep{gao2025principled}, and take LLaMA-3-8B-Instruct as a representative case. For consistency, training is conducted on the datasets released by SimPO. Since the original dataset provides only reward model scores and lacks explicit rating scores, we employ GPT-4o-mini to generate rating scores as a substitute. Specifically, the models LLaMA-3-Instruct-8B-DPO and LLaMA-3-Instruct-8B-SimPO are publicly available from the SimPO repository. In addition, we present two versions of MixDPO based on different difficulty metrics, namely GPT-4o-mini rating score and original reward score. Table~\ref{tab:instruct_version_results} (Top) shows that MixDPO consistently outperforms these baselines.

\paragraph{Exploring the impact of difficulty threshold $\tau$.}  To understand the sensitivity of the threshold, we explore some alternative values $\tau \in \{0.5, 1, 1.5\}$, which correspond to selecting approximately 6,000 (10\% of the dataset), 18,653 (25\%), and 30,653 (50\%) difficult pairs, respectively. By default, we set the threshold at $\tau=0.5$, corresponding to the minimum margin in score differences. Table~\ref{tab:eval_diff_difficulty_threshold} (Bottom) demonstrate that the performance degrades with increasing values of the threshold $\tau$, highlighting the importance of the difficult threshold.
Empirically, selecting the top 10\% most difficult pairs (i.e., $\tau=0.5$) yields the best results, suggesting that moderate exposure to difficult pairs strikes a favorable balance between informativeness and overfitting. 

\begin{table}[h]
\centering
\caption{\textbf{Top}: Performance comparison of baselines under stronger \underline{LLaMA3-8B-Instruct} base model. We follow SimPO’s training setup and data~\citep{meng2024simpo}; since it only provides reward scores, we then generate rating scores with GPT-4o-mini for consistency.
\textbf{Bottom}: The impact of different difficulty threshold $\tau$. Here, we explore three threshold settings based on the minimum score gap, i.e., $\tau= \{0.5, 1, 1.5\}$. }
\begin{minipage}[t]{1\linewidth}
\centering
\resizebox{\linewidth}{!}{
\begin{tabular}{c|cc}
\toprule
~ & \multicolumn{2}{c}{\cellcolor{blue!10} \textbf{Model: LLaMA3-8B-Instruct}} \\
\cmidrule(lr){2-3}
\textbf{Method} & \textbf{LC Win Rate (\%)} & \textbf{Win Rate (\%)} \\
\midrule
DPO & 19.53 & 20.25 \\
SimPO & 24.68 & 22.73 \\
SelectiveDPO & 3.25 & 1.12 \\
\midrule
\method~ (Orig. reward) & \underline{29.02} & \textbf{28.01} \\
\method~ (GPT-4o-mini) & \textbf{29.47} & 
\underline{25.59} \\
\bottomrule
\end{tabular}}
\label{tab:instruct_version_results}
\end{minipage}\hfill

\begin{minipage}[t]{1\linewidth}
\centering
\resizebox{\linewidth}{!}{
\begin{tabular}{c|cc}
\toprule
~ & \multicolumn{2}{c}{\textbf{AlpacaEval 2.0}} \\
\cmidrule(lr){2-3}
\textbf{Method} & \textbf{LC Win Rate (\%)} & \textbf{Win Rate (\%)} \\
\midrule
\text{Base-SFT}  & 3.73 & 10.19 \\
\text{vanilla DPO} & 9.37 & 16.77 \\
\midrule
\method~ ($\tau=1.5$) & 3.80 & 2.61 \\
\method~ ($\tau=1$)   & 6.48 & 25.47 \\
\method~ ($\tau=0.5$) & 14.42 & 36.65 \\
\bottomrule
\end{tabular}}
\label{tab:eval_diff_difficulty_threshold}
\end{minipage}
\end{table}

\section{Conclusion and Limitations}\label{sec:conclusion}

In this work, we are interested in exploring an alternative way to handle difficult pairs with small margins that may hinder alignment due to potential likelihood displacement. Instead of filtering them out in common practice, we develop a simple yet effective method, \method, which adaptively switches from DPO loss to standard SFT loss based on the difficulty of preference pairs, leveraging both clear and ambiguous signals to boost alignment performance. 
Specifically, we utilize the DPO loss for confident, easy pairs and the SFT loss for difficult pairs.
Extensive empirical experiments validate the effectiveness of our proposed approach.

 In contrast to likelihood-displacement mitigation methods that either i) propose a single general-purpose preference loss for all samples \citep{xiao2024cal, pang2024iterative} or ii) score and filter out difficult ones \citep{razin2024unintentional}, our work advocate a complementary perspective: reuse difficult pairs by optimizing an SFT objective. This highlights the importance of matching training signals to data difficulty, rather than enforcing a one-size-fits-all objective, opens a promising direction for leveraging difficult data in a principled way to improve alignment.

Nonetheless, we acknowledge several limitations of our approach. While the proposed method outperforms all baselines, there may exist more effective strategies for handling difficult pairs beyond using the SFT loss, which presents a promising direction for future research. Then, although designing a better difficulty metric is not our primary goal, it also remains a promising direction worth exploring.
Besides, our difficulty metric primarily relies on the original LLM rating scores, which may be noisy or inaccurate, as noted in prior work \citep{pang2024improving}. Such potential scoring errors may distort the estimation of preference difficulty, leading to suboptimal training dynamics and ultimately hindering the alignment performance of LLMs.
 However, this practical concern can be substantially mitigated through existing score curation techniques \citep{zhu2023unmasking, pang2024improving}, and developing more advanced approaches remains an important direction for future work.

\section*{Impact Statement}
This paper advances LLM preference alignment by proposing a difficulty-aware training strategy that better leverages small-margin (ambiguous) preference pairs. The method is intended to improve the stability and effectiveness of alignment training and does not introduce new capabilities beyond standard preference optimization and supervised fine-tuning. Potential risks are primarily inherited from the underlying preference data and any automated rating signals used in practice (e.g., biases or systematic errors), as well as the general possibility that improved alignment techniques could be misused with harmful objectives. Based on our assessment, we do not foresee additional societal concerns specific to this method beyond these well-known considerations.

\bibliographystyle{icml2026}
\bibliography{references}

\begin{thebibliography}{65}
\providecommand{\natexlab}[1]{#1}
\providecommand{\url}[1]{\texttt{#1}}
\expandafter\ifx\csname urlstyle\endcsname\relax
  \providecommand{\doi}[1]{doi: #1}\else
  \providecommand{\doi}{doi: \begingroup \urlstyle{rm}\Url}\fi

\bibitem[Achiam et~al.(2023)Achiam, Adler, Agarwal, Ahmad, Akkaya, Aleman, Almeida, Altenschmidt, Altman, Anadkat, et~al.]{achiam2023gpt}
Achiam, J., Adler, S., Agarwal, S., Ahmad, L., Akkaya, I., Aleman, F.~L., Almeida, D., Altenschmidt, J., Altman, S., Anadkat, S., et~al.
\newblock Gpt-4 technical report.
\newblock \emph{arXiv preprint arXiv:2303.08774}, 2023.

\bibitem[Avrahami et~al.(1997)Avrahami, Kareev, Bogot, Caspi, Dunaevsky, and Lerner]{avrahami1997teaching}
Avrahami, J., Kareev, Y., Bogot, Y., Caspi, R., Dunaevsky, S., and Lerner, S.
\newblock Teaching by examples: Implications for the process of category acquisition.
\newblock \emph{The Quarterly Journal of Experimental Psychology Section A}, 50\penalty0 (3):\penalty0 586--606, 1997.

\bibitem[Azar et~al.(2024)Azar, Guo, Piot, Munos, Rowland, Valko, and Calandriello]{azar2024general}
Azar, M.~G., Guo, Z.~D., Piot, B., Munos, R., Rowland, M., Valko, M., and Calandriello, D.
\newblock A general theoretical paradigm to understand learning from human preferences.
\newblock In \emph{International Conference on Artificial Intelligence and Statistics}, pp.\  4447--4455. PMLR, 2024.

\bibitem[Bai et~al.(2022)Bai, Jones, Ndousse, Askell, Chen, DasSarma, Drain, Fort, Ganguli, Henighan, et~al.]{bai2022training}
Bai, Y., Jones, A., Ndousse, K., Askell, A., Chen, A., DasSarma, N., Drain, D., Fort, S., Ganguli, D., Henighan, T., et~al.
\newblock Training a helpful and harmless assistant with reinforcement learning from human feedback.
\newblock \emph{arXiv preprint arXiv:2204.05862}, 2022.

\bibitem[Bengio et~al.(2009)Bengio, Louradour, Collobert, and Weston]{bengio2009curriculum}
Bengio, Y., Louradour, J., Collobert, R., and Weston, J.
\newblock Curriculum learning.
\newblock In \emph{Proceedings of the 26th annual international conference on machine learning}, pp.\  41--48, 2009.

\bibitem[Chen et~al.(2023{\natexlab{a}})Chen, Wang, Shah, Tao, Wei, Xie, Sugiyama, and Raj]{chen2023understanding}
Chen, H., Wang, J., Shah, A., Tao, R., Wei, H., Xie, X., Sugiyama, M., and Raj, B.
\newblock Understanding and mitigating the label noise in pre-training on downstream tasks.
\newblock \emph{arXiv preprint arXiv:2309.17002}, 2023{\natexlab{a}}.

\bibitem[Chen et~al.(2023{\natexlab{b}})Chen, Li, Yan, Wang, Gunaratna, Yadav, Tang, Srinivasan, Zhou, Huang, et~al.]{chen2023alpagasus}
Chen, L., Li, S., Yan, J., Wang, H., Gunaratna, K., Yadav, V., Tang, Z., Srinivasan, V., Zhou, T., Huang, H., et~al.
\newblock Alpagasus: Training a better alpaca with fewer data.
\newblock \emph{arXiv preprint arXiv:2307.08701}, 2023{\natexlab{b}}.

\bibitem[Chowdhury et~al.(2024)Chowdhury, Kini, and Natarajan]{chowdhury2024provably}
Chowdhury, S.~R., Kini, A., and Natarajan, N.
\newblock Provably robust dpo: Aligning language models with noisy feedback.
\newblock \emph{arXiv preprint arXiv:2403.00409}, 2024.

\bibitem[Clark et~al.(2018)Clark, Cowhey, Etzioni, Khot, Sabharwal, Schoenick, and Tafjord]{clark2018think}
Clark, P., Cowhey, I., Etzioni, O., Khot, T., Sabharwal, A., Schoenick, C., and Tafjord, O.
\newblock Think you have solved question answering? try arc, the ai2 reasoning challenge.
\newblock \emph{arXiv preprint arXiv:1803.05457}, 2018.

\bibitem[Cobbe et~al.(2021)Cobbe, Kosaraju, Bavarian, Chen, Jun, Kaiser, Plappert, Tworek, Hilton, Nakano, Hesse, and Schulman]{cobbe2021gsm8k}
Cobbe, K., Kosaraju, V., Bavarian, M., Chen, M., Jun, H., Kaiser, L., Plappert, M., Tworek, J., Hilton, J., Nakano, R., Hesse, C., and Schulman, J.
\newblock Training verifiers to solve math word problems.
\newblock \emph{arXiv preprint arXiv:2110.14168}, 2021.

\bibitem[Croitoru et~al.(2024{\natexlab{a}})Croitoru, Hondru, Ionescu, Sebe, and Shah]{croitoru2024curriculum}
Croitoru, F.-A., Hondru, V., Ionescu, R.~T., Sebe, N., and Shah, M.
\newblock Curriculum direct preference optimization for diffusion and consistency models.
\newblock \emph{arXiv preprint arXiv:2405.13637}, 2024{\natexlab{a}}.

\bibitem[Croitoru et~al.(2024{\natexlab{b}})Croitoru, Hondru, Ionescu, and Shah]{croitoru2024reverse}
Croitoru, F.-A., Hondru, V., Ionescu, R.~T., and Shah, M.
\newblock Reverse stable diffusion: What prompt was used to generate this image?
\newblock \emph{Computer Vision and Image Understanding}, 249:\penalty0 104210, 2024{\natexlab{b}}.

\bibitem[Cui et~al.()Cui, Yuan, Ding, Yao, Zhu, Ni, Xie, Liu, and Sun]{cuiultrafeedback}
Cui, G., Yuan, L., Ding, N., Yao, G., Zhu, W., Ni, Y., Xie, G., Liu, Z., and Sun, M.
\newblock Ultrafeedback: Boosting language models with high-quality feedback, 2024.
\newblock In \emph{URL https://openreview. net/forum}.

\bibitem[Cui et~al.(2023)Cui, Yuan, Ding, Yao, Zhu, Ni, Xie, Liu, and Sun]{cui2023ultrafeedback}
Cui, G., Yuan, L., Ding, N., Yao, G., Zhu, W., Ni, Y., Xie, G., Liu, Z., and Sun, M.
\newblock Ultrafeedback: Boosting language models with high-quality feedback.
\newblock 2023.

\bibitem[Deng et~al.(2025)Deng, Zhong, Ai, Feng, Wang, and He]{deng2025less}
Deng, X., Zhong, H., Ai, R., Feng, F., Wang, Z., and He, X.
\newblock Less is more: Improving llm alignment via preference data selection.
\newblock \emph{arXiv preprint arXiv:2502.14560}, 2025.

\bibitem[Ethayarajh et~al.()Ethayarajh, Choi, and Swayamdipta]{ethayarajh2110understanding}
Ethayarajh, K., Choi, Y., and Swayamdipta, S.
\newblock Understanding dataset difficulty with v-usable information (2021).
\newblock \emph{URL https://arxiv. org/abs/2110.08420}.

\bibitem[Ethayarajh et~al.(2024)Ethayarajh, Xu, Muennighoff, Jurafsky, and Kiela]{ethayarajh2024kto}
Ethayarajh, K., Xu, W., Muennighoff, N., Jurafsky, D., and Kiela, D.
\newblock Kto: Model alignment as prospect theoretic optimization.
\newblock \emph{arXiv preprint arXiv:2402.01306}, 2024.

\bibitem[Gao et~al.(2025)Gao, Li, Liu, Xie, Zhao, and Xu]{gao2025principled}
Gao, C., Li, H., Liu, L., Xie, Z., Zhao, P., and Xu, Z.
\newblock Principled data selection for alignment: The hidden risks of difficult examples.
\newblock \emph{arXiv preprint arXiv:2502.09650}, 2025.

\bibitem[Grattafiori et~al.(2024)Grattafiori, Dubey, Jauhri, Pandey, Kadian, Al-Dahle, Letman, Mathur, Schelten, Vaughan, et~al.]{grattafiori2024llama}
Grattafiori, A., Dubey, A., Jauhri, A., Pandey, A., Kadian, A., Al-Dahle, A., Letman, A., Mathur, A., Schelten, A., Vaughan, A., et~al.
\newblock The llama 3 herd of models.
\newblock \emph{arXiv preprint arXiv:2407.21783}, 2024.

\bibitem[Hendrycks et~al.(2020)Hendrycks, Burns, Basart, Zou, Mazeika, Song, and Steinhardt]{hendrycks2020measuring}
Hendrycks, D., Burns, C., Basart, S., Zou, A., Mazeika, M., Song, D., and Steinhardt, J.
\newblock Measuring massive multitask language understanding.
\newblock \emph{arXiv preprint arXiv:2009.03300}, 2020.

\bibitem[Hong et~al.(2024)Hong, Lee, and Thorne]{hong2024orpo}
Hong, J., Lee, N., and Thorne, J.
\newblock Orpo: Monolithic preference optimization without reference model.
\newblock \emph{arXiv preprint arXiv:2403.07691}, 2024.

\bibitem[Huang et~al.(2025)Huang, Wu, Chen, Wang, Gao, Ding, Wu, He, and Wang]{huang2025larger}
Huang, K., Wu, J., Chen, Z., Wang, X., Gao, J., Ding, B., Wu, J., He, X., and Wang, X.
\newblock Larger or smaller reward margins to select preferences for alignment?
\newblock \emph{arXiv preprint arXiv:2503.01864}, 2025.

\bibitem[Jiang et~al.(2023)Jiang, Sablayrolles, Mensch, Bamford, Chaplot, Casas, Bressand, Lengyel, Lample, Saulnier, et~al.]{jiang2023mistral}
Jiang, A.~Q., Sablayrolles, A., Mensch, A., Bamford, C., Chaplot, D.~S., Casas, D. d.~l., Bressand, F., Lengyel, G., Lample, G., Saulnier, L., et~al.
\newblock Mistral 7b.
\newblock \emph{arXiv preprint arXiv:2310.06825}, 2023.

\bibitem[Jiang et~al.(2014)Jiang, Meng, Mitamura, and Hauptmann]{jiang2014easy}
Jiang, L., Meng, D., Mitamura, T., and Hauptmann, A.~G.
\newblock Easy samples first: Self-paced reranking for zero-example multimedia search.
\newblock In \emph{Proceedings of the 22nd ACM international conference on Multimedia}, pp.\  547--556, 2014.

\bibitem[Kim \& Lee(2024)Kim and Lee]{kim2024strategic}
Kim, J. and Lee, J.
\newblock Strategic data ordering: Enhancing large language model performance through curriculum learning.
\newblock \emph{arXiv preprint arXiv:2405.07490}, 2024.

\bibitem[Kong et~al.()Kong, Xu, Wang, Zhang, and Kankanhalli]{kongperplexity}
Kong, K., Xu, X., Wang, D., Zhang, J., and Kankanhalli, M.
\newblock Perplexity-aware correction for robust alignment with noisy preferences.
\newblock In \emph{The Thirty-eighth Annual Conference on Neural Information Processing Systems}.

\bibitem[Li et~al.(2024)Li, Chiang, Frick, Dunlap, Zhu, Gonzalez, and Stoica]{li2024live}
Li, T., Chiang, W.-L., Frick, E., Dunlap, L., Zhu, B., Gonzalez, J.~E., and Stoica, I.
\newblock From live data to high-quality benchmarks: The arena-hard pipeline, 2024.

\bibitem[Li et~al.(2023)Li, Zhang, Dubois, Taori, Gulrajani, Guestrin, Liang, and Hashimoto]{li2023alpacaeval}
Li, X., Zhang, T., Dubois, Y., Taori, R., Gulrajani, I., Guestrin, C., Liang, P., and Hashimoto, T.~B.
\newblock Alpacaeval: An automatic evaluator of instruction-following models, 2023.

\bibitem[Lin et~al.(2021)Lin, Hilton, and Evans]{lin2021truthfulqa}
Lin, S., Hilton, J., and Evans, O.
\newblock Truthfulqa: Measuring how models mimic human falsehoods.
\newblock \emph{arXiv preprint arXiv:2109.07958}, 2021.

\bibitem[Lin et~al.(2024)Lin, Gou, Gong, Liu, Shen, Xu, Lin, Yang, Jiao, Duan, et~al.]{lin2024rho}
Lin, Z., Gou, Z., Gong, Y., Liu, X., Shen, Y., Xu, R., Lin, C., Yang, Y., Jiao, J., Duan, N., et~al.
\newblock Rho-1: Not all tokens are what you need.
\newblock \emph{arXiv preprint arXiv:2404.07965}, 2024.

\bibitem[Liu et~al.(2024)Liu, Di, Wei, Wang, Zhang, Xiao, Wang, Pang, Chen, Shah, et~al.]{liu2024automatic}
Liu, M., Di, Z., Wei, J., Wang, Z., Zhang, H., Xiao, R., Wang, H., Pang, J., Chen, H., Shah, A., et~al.
\newblock Automatic dataset construction (adc): Sample collection, data curation, and beyond.
\newblock \emph{arXiv preprint arXiv:2408.11338}, 2024.

\bibitem[Liu et~al.(2023)Liu, Zeng, He, Jiang, and He]{liu2023makes}
Liu, W., Zeng, W., He, K., Jiang, Y., and He, J.
\newblock What makes good data for alignment? a comprehensive study of automatic data selection in instruction tuning.
\newblock \emph{arXiv preprint arXiv:2312.15685}, 2023.

\bibitem[Liu \& Guo(2020)Liu and Guo]{liu2020peer}
Liu, Y. and Guo, H.
\newblock Peer loss functions: Learning from noisy labels without knowing noise rates.
\newblock In \emph{International conference on machine learning}, pp.\  6226--6236. PMLR, 2020.

\bibitem[Meng et~al.(2024)Meng, Xia, and Chen]{meng2024simpo}
Meng, Y., Xia, M., and Chen, D.
\newblock Simpo: Simple preference optimization with a reference-free reward.
\newblock \emph{arXiv preprint arXiv:2405.14734}, 2024.

\bibitem[Mitchell()]{mitchellnote}
Mitchell, E.
\newblock A note on dpo with noisy preferences and relationship to ipo, 2023.
\newblock \emph{URL https://ericmitchell. ai/cdpo. pdf}.

\bibitem[Nakano et~al.(2021)Nakano, Hilton, Balaji, Wu, Ouyang, Kim, Hesse, Jain, Kosaraju, Saunders, et~al.]{nakano2021webgpt}
Nakano, R., Hilton, J., Balaji, S., Wu, J., Ouyang, L., Kim, C., Hesse, C., Jain, S., Kosaraju, V., Saunders, W., et~al.
\newblock Webgpt: Browser-assisted question-answering with human feedback.
\newblock \emph{arXiv preprint arXiv:2112.09332}, 2021.

\bibitem[Natarajan et~al.(2013)Natarajan, Dhillon, Ravikumar, and Tewari]{natarajan2013learning}
Natarajan, N., Dhillon, I.~S., Ravikumar, P.~K., and Tewari, A.
\newblock Learning with noisy labels.
\newblock \emph{Advances in neural information processing systems}, 26, 2013.

\bibitem[Ouyang et~al.(2022)Ouyang, Wu, Jiang, Almeida, Wainwright, Mishkin, Zhang, Agarwal, Slama, Ray, et~al.]{ouyang2022training}
Ouyang, L., Wu, J., Jiang, X., Almeida, D., Wainwright, C., Mishkin, P., Zhang, C., Agarwal, S., Slama, K., Ray, A., et~al.
\newblock Training language models to follow instructions with human feedback.
\newblock \emph{Advances in neural information processing systems}, 35:\penalty0 27730--27744, 2022.

\bibitem[Pal et~al.(2024)Pal, Karkhanis, Dooley, Roberts, Naidu, and White]{pal2024smaug}
Pal, A., Karkhanis, D., Dooley, S., Roberts, M., Naidu, S., and White, C.
\newblock Smaug: Fixing failure modes of preference optimisation with dpo-positive.
\newblock \emph{arXiv preprint arXiv:2402.13228}, 2024.

\bibitem[Pang et~al.(2024{\natexlab{a}})Pang, Wang, Zhu, Yao, Qian, and Liu]{pang2024fairness}
Pang, J., Wang, J., Zhu, Z., Yao, Y., Qian, C., and Liu, Y.
\newblock Fairness without harm: An influence-guided active sampling approach.
\newblock \emph{Advances in Neural Information Processing Systems}, 37:\penalty0 61513--61548, 2024{\natexlab{a}}.

\bibitem[Pang et~al.(2024{\natexlab{b}})Pang, Wei, Shah, Zhu, Wang, Qian, Liu, Bao, and Wei]{pang2024improving}
Pang, J., Wei, J., Shah, A.~P., Zhu, Z., Wang, Y., Qian, C., Liu, Y., Bao, Y., and Wei, W.
\newblock Improving data efficiency via curating llm-driven rating systems.
\newblock \emph{arXiv preprint arXiv:2410.10877}, 2024{\natexlab{b}}.

\bibitem[Pang et~al.(2025)Pang, Di, Zhu, Wei, Cheng, Qian, and Liu]{pang2025token}
Pang, J., Di, N., Zhu, Z., Wei, J., Cheng, H., Qian, C., and Liu, Y.
\newblock Token cleaning: Fine-grained data selection for llm supervised fine-tuning.
\newblock \emph{arXiv preprint arXiv:2502.01968}, 2025.

\bibitem[Pang et~al.(2024{\natexlab{c}})Pang, Yuan, He, Cho, Sukhbaatar, and Weston]{pang2024iterative}
Pang, R.~Y., Yuan, W., He, H., Cho, K., Sukhbaatar, S., and Weston, J.
\newblock Iterative reasoning preference optimization.
\newblock \emph{Advances in Neural Information Processing Systems}, 37:\penalty0 116617--116637, 2024{\natexlab{c}}.

\bibitem[Park et~al.(2024)Park, Rafailov, Ermon, and Finn]{park2024disentangling}
Park, R., Rafailov, R., Ermon, S., and Finn, C.
\newblock Disentangling length from quality in direct preference optimization.
\newblock \emph{arXiv preprint arXiv:2403.19159}, 2024.

\bibitem[Platanios et~al.(2019)Platanios, Stretcu, Neubig, Poczos, and Mitchell]{platanios2019competence}
Platanios, E.~A., Stretcu, O., Neubig, G., Poczos, B., and Mitchell, T.~M.
\newblock Competence-based curriculum learning for neural machine translation.
\newblock \emph{arXiv preprint arXiv:1903.09848}, 2019.

\bibitem[Rafailov et~al.(2024{\natexlab{a}})Rafailov, Hejna, Park, and Finn]{rafailov2024r}
Rafailov, R., Hejna, J., Park, R., and Finn, C.
\newblock From $ r $ to $ q^* $: Your language model is secretly a q-function.
\newblock \emph{arXiv preprint arXiv:2404.12358}, 2024{\natexlab{a}}.

\bibitem[Rafailov et~al.(2024{\natexlab{b}})Rafailov, Sharma, Mitchell, Manning, Ermon, and Finn]{rafailov2024direct}
Rafailov, R., Sharma, A., Mitchell, E., Manning, C.~D., Ermon, S., and Finn, C.
\newblock Direct preference optimization: Your language model is secretly a reward model.
\newblock \emph{Advances in Neural Information Processing Systems}, 36, 2024{\natexlab{b}}.

\bibitem[Razin et~al.(2024)Razin, Malladi, Bhaskar, Chen, Arora, and Hanin]{razin2024unintentional}
Razin, N., Malladi, S., Bhaskar, A., Chen, D., Arora, S., and Hanin, B.
\newblock Unintentional unalignment: Likelihood displacement in direct preference optimization.
\newblock \emph{arXiv preprint arXiv:2410.08847}, 2024.

\bibitem[Sakaguchi et~al.(2021)Sakaguchi, Bras, Bhagavatula, and Choi]{sakaguchi2021winogrande}
Sakaguchi, K., Bras, R.~L., Bhagavatula, C., and Choi, Y.
\newblock Winogrande: An adversarial winograd schema challenge at scale.
\newblock \emph{Communications of the ACM}, 64\penalty0 (9):\penalty0 99--106, 2021.

\bibitem[Stiennon et~al.(2020)Stiennon, Ouyang, Wu, Ziegler, Lowe, Voss, Radford, Amodei, and Christiano]{stiennon2020learning}
Stiennon, N., Ouyang, L., Wu, J., Ziegler, D., Lowe, R., Voss, C., Radford, A., Amodei, D., and Christiano, P.~F.
\newblock Learning to summarize with human feedback.
\newblock \emph{Advances in neural information processing systems}, 33:\penalty0 3008--3021, 2020.

\bibitem[Tudor~Ionescu et~al.(2016)Tudor~Ionescu, Alexe, Leordeanu, Popescu, Papadopoulos, and Ferrari]{tudor2016hard}
Tudor~Ionescu, R., Alexe, B., Leordeanu, M., Popescu, M., Papadopoulos, D.~P., and Ferrari, V.
\newblock How hard can it be? estimating the difficulty of visual search in an image.
\newblock In \emph{Proceedings of the IEEE Conference on Computer Vision and Pattern Recognition}, pp.\  2157--2166, 2016.

\bibitem[Wu et~al.(2024)Wu, Meng, and Chen]{wu2024curriculum}
Wu, B., Meng, F., and Chen, L.
\newblock Curriculum learning with quality-driven data selection.
\newblock \emph{arXiv preprint arXiv:2407.00102}, 2024.

\bibitem[Xia et~al.(2024)Xia, Malladi, Gururangan, Arora, and Chen]{xia2024less}
Xia, M., Malladi, S., Gururangan, S., Arora, S., and Chen, D.
\newblock Less: Selecting influential data for targeted instruction tuning.
\newblock \emph{arXiv preprint arXiv:2402.04333}, 2024.

\bibitem[Xia et~al.(2020)Xia, Liu, Han, Wang, Gong, Liu, Niu, Tao, and Sugiyama]{xia2020part}
Xia, X., Liu, T., Han, B., Wang, N., Gong, M., Liu, H., Niu, G., Tao, D., and Sugiyama, M.
\newblock Part-dependent label noise: Towards instance-dependent label noise.
\newblock \emph{Advances in neural information processing systems}, 33:\penalty0 7597--7610, 2020.

\bibitem[Xiao et~al.(2024)Xiao, Yuan, Zhu, Li, and Honavar]{xiao2024cal}
Xiao, T., Yuan, Y., Zhu, H., Li, M., and Honavar, V.~G.
\newblock Cal-dpo: Calibrated direct preference optimization for language model alignment.
\newblock \emph{Advances in Neural Information Processing Systems}, 37:\penalty0 114289--114320, 2024.

\bibitem[Xu et~al.(2024{\natexlab{a}})Xu, Sharaf, Chen, Tan, Shen, Van~Durme, Murray, and Kim]{xu2024contrastive}
Xu, H., Sharaf, A., Chen, Y., Tan, W., Shen, L., Van~Durme, B., Murray, K., and Kim, Y.~J.
\newblock Contrastive preference optimization: Pushing the boundaries of llm performance in machine translation.
\newblock \emph{arXiv preprint arXiv:2401.08417}, 2024{\natexlab{a}}.

\bibitem[Xu et~al.(2024{\natexlab{b}})Xu, Fu, Gao, Ye, Liu, Mei, Wang, Yu, and Wu]{xu2024dpo}
Xu, S., Fu, W., Gao, J., Ye, W., Liu, W., Mei, Z., Wang, G., Yu, C., and Wu, Y.
\newblock Is dpo superior to ppo for llm alignment? a comprehensive study.
\newblock \emph{arXiv preprint arXiv:2404.10719}, 2024{\natexlab{b}}.

\bibitem[Yang et~al.(2024)Yang, Yang, Zhang, Hui, Zheng, Yu, Li, Liu, Huang, Wei, et~al.]{yang2024qwen2}
Yang, A., Yang, B., Zhang, B., Hui, B., Zheng, B., Yu, B., Li, C., Liu, D., Huang, F., Wei, H., et~al.
\newblock Qwen2. 5 technical report.
\newblock \emph{arXiv preprint arXiv:2412.15115}, 2024.

\bibitem[Yang et~al.(2025)Yang, Li, Di, Pang, Zhou, Cheng, Han, and Wei]{yang2025entp}
Yang, Z., Li, L., Di, N., Pang, J., Zhou, Y., Cheng, H., Han, B., and Wei, J.
\newblock Entp: Enhancing low-quality sft data via neural-symbolic text purge-mix.
\newblock \emph{arXiv preprint arXiv:2510.23160}, 2025.

\bibitem[Yuan et~al.(2024)Yuan, Cui, Wang, Ding, Wang, Deng, Shan, Chen, Xie, Lin, et~al.]{yuan2024advancing}
Yuan, L., Cui, G., Wang, H., Ding, N., Wang, X., Deng, J., Shan, B., Chen, H., Xie, R., Lin, Y., et~al.
\newblock Advancing llm reasoning generalists with preference trees.
\newblock \emph{arXiv preprint arXiv:2404.02078}, 2024.

\bibitem[Zellers et~al.(2019)Zellers, Holtzman, Bisk, Farhadi, and Choi]{zellers2019hellaswag}
Zellers, R., Holtzman, A., Bisk, Y., Farhadi, A., and Choi, Y.
\newblock Hellaswag: Can a machine really finish your sentence?
\newblock \emph{arXiv preprint arXiv:1905.07830}, 2019.

\bibitem[Zhang et~al.(2025)Zhang, Pang, Zhu, and Liu]{zhang2025evaluating}
Zhang, Y., Pang, J., Zhu, Z., and Liu, Y.
\newblock Evaluating llm-corrupted crowdsourcing data without ground truth.
\newblock \emph{arXiv preprint arXiv:2506.06991}, 2025.

\bibitem[Zhao et~al.(2023)Zhao, Joshi, Liu, Khalman, Saleh, and Liu]{zhao2023slic}
Zhao, Y., Joshi, R., Liu, T., Khalman, M., Saleh, M., and Liu, P.~J.
\newblock Slic-hf: Sequence likelihood calibration with human feedback.
\newblock \emph{arXiv preprint arXiv:2305.10425}, 2023.

\bibitem[Zheng et~al.(2023)Zheng, Chiang, Sheng, Zhuang, Wu, Zhuang, Lin, Li, Li, Xing, et~al.]{zheng2023judging}
Zheng, L., Chiang, W.-L., Sheng, Y., Zhuang, S., Wu, Z., Zhuang, Y., Lin, Z., Li, Z., Li, D., Xing, E., et~al.
\newblock Judging llm-as-a-judge with mt-bench and chatbot arena.
\newblock \emph{Advances in Neural Information Processing Systems}, 36:\penalty0 46595--46623, 2023.

\bibitem[Zhu et~al.(2023)Zhu, Wang, Cheng, and Liu]{zhu2023unmasking}
Zhu, Z., Wang, J., Cheng, H., and Liu, Y.
\newblock Unmasking and improving data credibility: A study with datasets for training harmless language models.
\newblock \emph{arXiv preprint arXiv:2311.11202}, 2023.

\end{thebibliography}

\newpage
\appendix
\onecolumn

\section*{Appendix}\label{sec:appendix}

\section*{Organization of the Appendix}
\label{sec:organization_of_appendix}

\squishlist
    \item \textbf{Section~\ref{sec:apx_experiment_details}}: Describes the experimental setup of this work, including the computing environment, the SFT base models used, and their corresponding hyperparameters.
    \begin{itemize}
        \item Sec.~\ref{sec:computation_env}: Computing environment description;
        \item Sec.~\ref{sec:sft_base_model}: Used SFT base models;
        \item Sec.~\ref{sec:training_details}: Training details of preference alignment.
    \end{itemize}
    \item \textbf{Section~\ref{sec:more_experiment_results}}: Presents additional experimental results:
\begin{itemize}
    \item Sec.~\ref{sec:impact_of_lr}: The impact of learning rate;
    \item Sec.~\ref{sec:impact_of_difficulty_metrics}: Exploring the impact of alternative difficulty metrics;
    \item Sec.~\ref{apx:compare_with_likelihood_displacement_method}: Comparisons with alternative likelihood displacement methods;
    \item Sec.~\ref{sec:impact_of_absolute_quality}: The effect of absolute quality in preference pairs;
    \item Sec.~\ref{apx:sft_variants}: Explore two potential SFT-style baselines: 1) full SFT training on positive (chosen) responses, 2) full SFT training on both positive (chosen) and negative (rejected) responses.
    \item Sec.~\ref{sec:impact_of_score_noise}: The robustness of MixDPO under score noise;
    \item Sec.~\ref{sec:downstream_task_evaluation}: Downstream task evaluation.
\end{itemize}
\textbf{Section~\ref{apx:difficult_pairs_examples}}: Provides illustrative examples of easy and difficult preference pairs identified from the original rating scores.

\squishend

\section{Experimental Details}\label{sec:apx_experiment_details}

\subsection{Computation Environment}
\label{sec:computation_env}
In this work, all experiments were conducted on a server equipped with 8×NVIDIA L40S GPUs, each with 45 GB of memory. For full-parameter fine-tuning, reproducing each training run on 7B/8B models takes approximately 3 GPU hours.

\subsection{SFT Base Models}
\label{sec:sft_base_model}
In this work, we perform preference optimization experiments using publicly available SFT models that follow the standard post-training pipeline. Both models were fine-tuned on the Ultrachat 200k dataset. The publicly released hyperparameters used during the SFT training process are summarized in Table~\ref{tab:sft_hyperparameter_setting}. 

\begin{table}[h]
    \centering
    \caption{Training details for SFT models used in this work.}
     \resizebox{\linewidth}{!}{
    \begin{tabular}{c|cccccc}
    \toprule
    \textbf{SFT Model} & \textbf{Hugginface Source}  & \textbf{Batch Size} & \textbf{Learning Rate} & \textbf{Optimizer} & \textbf{LoRA?} \\
    \midrule
         LLaMA-3-8B-SFT & princeton-nlp/Llama-3-Base-8B-SFT  & 128 & 2e-5 & Adam & No \\
         Mistral-7B-SFT & HuggingFaceH4/mistral-7b-sft-beta & 128 & 2e-5 & Adam & No\\
         Qwen-2.5-7B-SFT & AmberYifan/Qwen2.5-7B-sft-ultrachat & 128 & 1e-5 & Adam & No \\
    \bottomrule
    \end{tabular}}
    \label{tab:sft_hyperparameter_setting}
\end{table}

\subsection{Preference Alignment Training Details}
\label{sec:training_details}

Table~\ref{tab:dpo_hyperparameter_setting} summarizes several key hyperparameters used in \method~ for our experiments. By default, we use a cosine learning rate scheduler. Following setting of SimPO \citep{meng2024simpo}, we set the maximum prompt length to 512 and the maximum sequence length to 1024. Additionally, all models are fine-tuned using BF16 precision. For baselines, we adopt the released models from SimPO \citep{meng2024simpo}, where the corresponding hyperparameters are provided.

\begin{table}[h]
    \centering
    \caption{Key hyper-parameters in \method~used for experiments.}
     \resizebox{0.8\linewidth}{!}{
    \begin{tabular}{c|cccccc}
    \toprule
\textbf{SFT model} & \textbf{Learning Rate} & \textbf{Batch Size} & \textbf{$\beta$} & \textbf{Epoch} & \textbf{Optimizer} & \textbf{LoRA?} \\
\midrule
     LLaMA-3-8B-SFT & 1e-6  & 128 & 0.01 & 1 & adamw\_torch & No \\
     Mistral-7B-SFT & 2e-7 & 128 & 0.01 & 1&adamw\_torch & No \\
    Qwen-2.7-7B & 8e-7 & 128 & 0.01 & 1&adamw\_torch & No \\
    \bottomrule
    \end{tabular}}
    \label{tab:dpo_hyperparameter_setting}
\end{table}

\paragraph{Additional base model Qwen-2.5-7B}  For the additional base model Qwen-2.5-7B, we evaluate several learning rate settings, and the corresponding results are presented in Table~\ref{tab:qwen_results} (Left). The learning rate for Qwen-2.5-7B is adopted from SelectiveDPO \citep{gao2025principled}.

\begin{table}[h]
    \centering
    \caption{Key hyper-parameters used for experiments in the Qwen-2.5-7B base model.}
     \resizebox{0.8\linewidth}{!}{
    \begin{tabular}{c|cccccc}
    \toprule
\textbf{Baseline} & \textbf{Learning Rate} & \textbf{Batch Size} & \textbf{$\beta$} & \textbf{Epoch} & \textbf{Optimizer} & \textbf{LoRA?} \\
\midrule
     DPO & 8e-7  & 128 & 0.01 & 1 & adamw\_torch & No \\
     SimPO & 8e-7 & 128 & 2.0 & 1 & adamw\_torch & No \\
    SelectiveDPO & 8e-7 & 128 & 0.01 & 1&adamw\_torch & No \\
    \method & 8e-7 & 128 & 0.01 & 1 &adamw\_torch & No \\
    \bottomrule
    \end{tabular}
    }
    \label{tab:qwen-hyper-parameter}
\end{table}

\paragraph{Additional preference dataset Argilla-7k} For a series of experiments on the Argilla-7k dataset,  we have examined several learning rate settings, whose results are provided in Table~\ref{tab:qwen_results} (Right). 
\begin{table}[h]
    \centering
    \caption{Key hyper-parameters used for experiments on the Argilla-7k preference dataset. We highlight the learning rate with the best result in \textbf{bold}.}
     \resizebox{0.85\linewidth}{!}{
    \begin{tabular}{c|cccccc}
    \toprule
\textbf{Baseline} & \textbf{Learning Rate Range} & \textbf{Batch Size} & \textbf{$\beta$} & \textbf{Epoch} & \textbf{Optimizer} & \textbf{LoRA?} \\
\midrule
     DPO & \{1e-7 5e-7 8e-7 \textbf{1e-6} 2e-6\} & 128 & 0.01 & 2 & adamw\_torch & No \\
     SimPO &  \{1e-7 5e-7 8e-7 \textbf{1e-6} 2e-6\}& 128 & 2.0 & 2 & adamw\_torch & No \\
    SelectiveDPO &  \{1e-7 5e-7 8e-7 \textbf{1e-6} 2e-6\} & 128 & 0.01 & 2&adamw\_torch & No \\
    \method &  \{1e-7 5e-7 8e-7 \textbf{1e-6} 2e-6\} & 128 & 0.01 & 2 &adamw\_torch & No \\

    \bottomrule
    \end{tabular}
    }
    \label{tab:argilla-hyper-parameter}
\end{table}

\section{More Experimental Results}\label{sec:more_experiment_results}

\subsection{Impact of Learning Rate}\label{sec:impact_of_lr}
We report the performance of our proposed method, \method, under different learning rates across two base models in Figure~\ref{fig:impact_of_lr}. For LLaMA-3-8B, the best performance is achieved with a learning rate of 1e-6, while for Mistral-7B, the optimal learning rate is 2e-7.

\begin{figure}[ht]
    \centering
    \begin{minipage}[t]{0.45\linewidth}
        \centering
    \includegraphics[width=\linewidth]{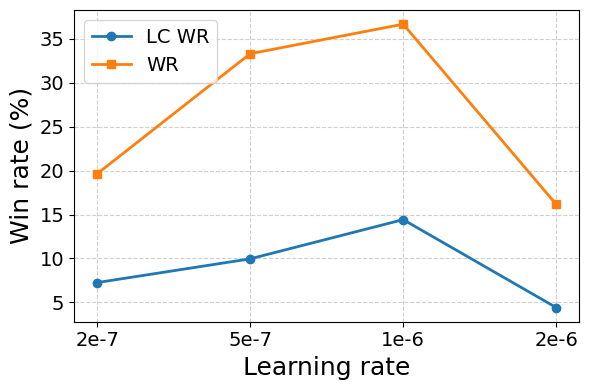}
        \subcaption{LLaMA-3-8B}
        \label{fig:impact_of_lr}
    \end{minipage}
    \hfill
    \begin{minipage}[t]{0.45\linewidth}
        \centering
    \includegraphics[width=\linewidth]{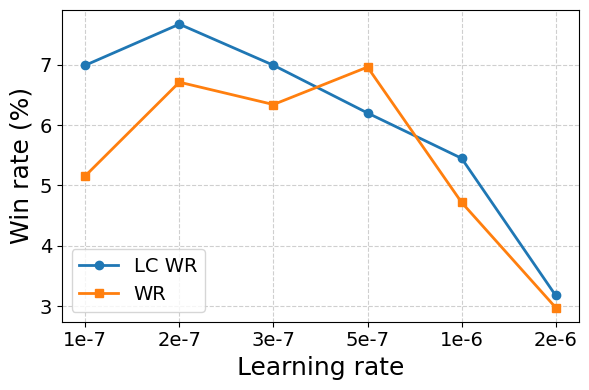}
        \subcaption{Mistral-7B}
        \label{fig:impact_of_lr}
    \end{minipage}
    \caption{Effect of learning rate across two base model configurations on the Ultrafeedback dataset.}
    \label{fig:impact_of_lr}
\end{figure}

\subsection{Exploring the Impact of Alternative Difficulty Metrics}
\label{sec:impact_of_difficulty_metrics}

\begin{figure}[t]
    \centering
    \hspace{-0.25in}
        \begin{minipage}[b]{0.33\linewidth}
    \centering
    \includegraphics[width=1\linewidth]{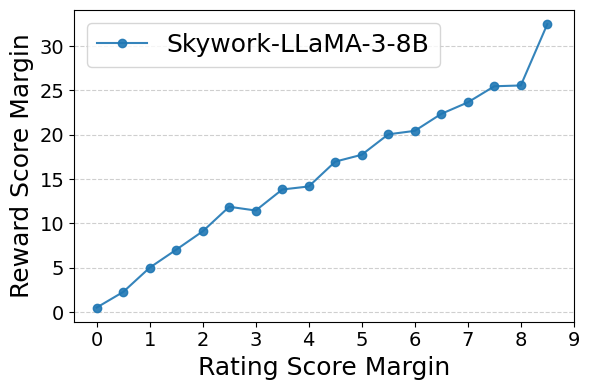}
    \subcaption{Reward score margin}
    \label{fig:enter-label}
    \end{minipage}%
    \begin{minipage}[b]{0.33\linewidth}
        \centering
    \includegraphics[width=1\linewidth]{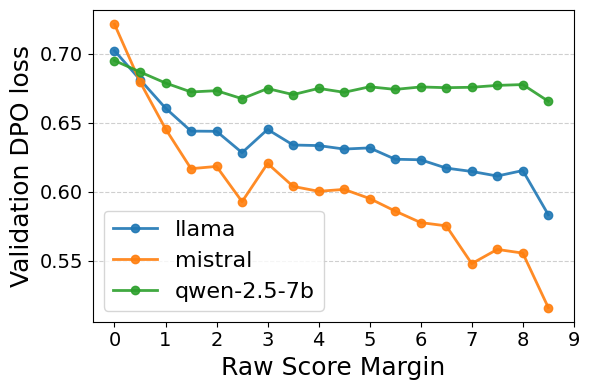}
        \subcaption{Validation DPO loss}
    \end{minipage}%
    \begin{minipage}[b]{0.33\linewidth}
        \centering
    \includegraphics[width=1\linewidth]{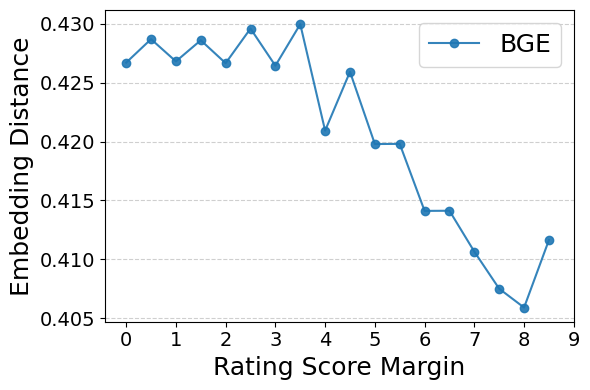}
        \subcaption{Embedding distance}
    \end{minipage}%
    \caption{Relationship between rating score margin and three alternative difficult metrics. Observe that the rating score margin correlates positively with reward score and DPO loss, but shows a counterintuitive negative correlation with embedding distance.}
    \label{fig:difficult_metric_relationship}
\end{figure}

\paragraph{Correlation between rating score margin and other difficulty metrics}
Here, we first systematically investigate the relationship between rating score margin and other alternative difficult metrics, including validation DPO loss \citep{gao2025principled}, reward score \citep{croitoru2024curriculum}, and embedding distance between responses. 
For the reward score, we utilize a powerful reward model from the RewardBench leaderboard, Skywork-Reward-Llama-3.1-8B\footnote{\url{https://huggingface.co/Skywork/Skywork-Reward-Llama-3.1-8B-v0.2}}.
To compute the embedding distance, we select the newly released open-source model,  BGE\footnote{\href{https://huggingface.co/BAAI/bge-large-en-v1.5}{\texttt{BAAI/bge-large-en-v1.5}}} as the embedding model. Note that the validation DPO losses are taken from SelectiveDPO and are based on three different base models, including Qwen-2.5-7B, Mistral-7B, and LLaMA-3-8B. Figure~\ref{fig:difficult_metric_relationship} shows a clear positive correlation between the raw score margin and both the reward score margin and validation DPO loss. A smaller validation DPO loss indicates that the sample is easier for the model to learn.
In contrast, a counter-intuitive negative correlation is observed between the rating score margin and embedding distance, which may be attributed to the representation limitations of the embedding model.
Notably, the original LLM rating score proves to be both sufficient and competitive for assessing sample-level difficulty, offering a highly efficient alternative to more costly methods such as reward model inference or validation DPO loss computation.

To evaluate the generalization of MixDPO under alternative difficulty notions, we investigate the embedding distance between preferred and dispreferred responses as well as the reward-score margin.
 Table~\ref{tab:eval_diff_difficulty_metrics} reports the alignment performance under each metric, highlighting the effectiveness of the original LLM rating score. Notably, the LLM rating score requires no additional computation, making it a highly efficient choice compared to alternatives that rely on reward model inference or embedding-based similarity.
Here, the reward score margin metric yields worse performance, primarily due to the limitations of the reward model used. The learning rates used for reward score margin and embedding distance are both 1e-7.

\begin{table}[ht]
    \caption{Comparison of alignment performance under different difficulty metrics.}
    \centering
        \resizebox{0.65\linewidth}{!}{
    \begin{tabular}{c|cc}
    \toprule
~ & \multicolumn{2}{c}{\textbf{AlpacaEval 2.0}} \\
\cmidrule{2-3}
\textbf{Method} & \textbf{LC Win Rate (\%)} & \textbf{Win Rate (\%)} \\
\midrule
\text{Base-SFT}  & 3.73    & 10.19 \\
\text{vanilla DPO} & 9.37 & 16.77 \\
\text{SimPO} & 6.77 & 14.04 \\
\midrule
MixDPO (Embedding Distance) &        8.75       &       19.13     \\ %
MixDPO (Reward Score Margin) &7.05      &        27.95 \\ %
\midrule
MixDPO (Orig. Rating Score) &  14.42 & 36.65  \\ 
\bottomrule
\end{tabular}}
\label{tab:eval_diff_difficulty_metrics}
\end{table}

\subsection{Comparison with Alternative Likelihood Displacement Methods}\label{apx:compare_with_likelihood_displacement_method}

Here, we investigate whether existing approaches for addressing likelihood displacement can be leveraged to handle difficult preference pairs \citep{razin2024unintentional, xiao2024cal, pang2024iterative}. A large proportion of these methods focus on adjustments to the loss function.
For instance, Smaug~\citep{pal2024smaug} introduces the DPOP loss function, which incorporates an additional term into the standard DPO loss. For clarity and completeness, we reproduce the DPOP loss function as follows.

\begin{equation}
\begin{aligned}
\mathcal{L}_{\text{DPOP}}(\pi_\theta; \pi_{\text{ref}}) 
= - \mathbb{E}_{(x, y_w, y_l) \sim D} \Bigg[ 
 \log \sigma \Bigg( 
    \beta \Big( 
        \log \frac{\pi_\theta(y_w \mid x)}{\pi_{\text{ref}}(y_w \mid x)} 
        - \log \frac{\pi_\theta(y_l \mid x)}{\pi_{\text{ref}}(y_l \mid x)} \underbrace{ - \lambda \cdot \max \left( 
    0, 
    \log \frac{\pi_{\text{ref}}(y_w \mid x)}{\pi_\theta(y_w \mid x)} 
\right)}_{\text{Additional term}}
\Big) 
\Bigg) 
\Bigg]
\end{aligned}
\tag{3}
\end{equation}

For CHES \citep{razin2024unintentional}, which uses the CHES score to filter out samples, we computed CHES scores on the UltraFeedback dataset and filtered out a specified proportion of samples accordingly. For Cal-DPO \citep{xiao2024cal}, we directly applied its proposed loss function during training and evaluated the resulting performance. Additionally, we implemented another recently proposed method that addresses the likelihood displacement problem by incorporating a negative log-likelihood (NLL) loss term into the DPO loss \citep{pang2024iterative}. We refer to this variant as \textbf{DPO+NLL}.

For CHES, similar to SelectiveDPO \citep{gao2025principled} and MixDPO (a special case), we experiment with two selection proportions of data samples: 50\% and 90\%. As shown in Table~\ref{tab:likelihood_displacement}, MixDPO consistently outperforms prior approaches aimed at mitigating the unintentional likelihood problem, demonstrating the effectiveness of our approach. 

\begin{table}[ht]
\caption{Comparison of alignment performance across different methods for mitigating likelihood displacement. The preference dataset is Ultrafeedback, and the base model used is LLaMA-3-8B.}
\centering
\resizebox{0.75\linewidth}{!}{
\begin{tabular}{l|cc}
    \toprule
~ & \multicolumn{2}{c}{\textbf{AlpacaEval 2.0}} \\
\cmidrule{2-3}
\textbf{Method} & \textbf{LC Win Rate (\%)} & \textbf{Win Rate (\%)} \\
\midrule
\text{Base-SFT}  & 3.73    & 10.19 \\
    \text{vanilla DPO} & 9.37 & 16.77 \\
\midrule
Cal-DPO \citep{xiao2024cal} & 4.56 & 7.45 \\
DPO + NLL \citep{pang2024iterative} & 4.25 & 8.45 \\
DPOP \citep{pal2024smaug} &     4.53   &   4.04      \\
CHES (Selected prop: 50\%) \citep{razin2024unintentional}	&6.16	& 11.93 \\
CHES (Selected prop: 90\%) &	8.13	& 13.91 \\
\method & 14.42 & 36.65 \\ 
\bottomrule
    \end{tabular}}
    \label{tab:likelihood_displacement}
\end{table}

\paragraph{Computational Overhead Comparison}
While our method adopts the curriculum learning framework, as in SelectiveDPO, it primarily challenges the conventional practice of filtering out difficult samples or those that cause likelihood displacement. MixDPO takes a fundamentally different approach by leveraging these difficult samples during the SFT phase to further enhance alignment performance. Empirical results validate the effectiveness of this strategy. Moreover, MixDPO introduces no additional computational overhead for computing difficulty metrics. To highlight this advantage, we report detailed overhead statistics in Table~\ref{tab:computational_overhead} using the UltraFeedback dataset with 60,000 samples. Specifically, computing DPO losses (SelectiveDPO) and CHES scores are performed on 8×H100 GPUs, taking approximately 20 minutes and 50 minutes, respectively. In comparison, training the full dataset on 8×H100 GPUs takes about 40 minutes. This demonstrates that difficulty-based filtering incurs non-trivial overhead, which our method successfully avoids—representing a key strength of MixDPO.

\begin{table}[h]
\centering
\caption{Introduced computational overhead (time) comparison on the Ultrafeedback dataset.}
\begin{tabular}{l c}
\toprule
\textbf{Metrics} & \textbf{Introduced Computational Overhead} \\
\midrule
CHES score \citep{razin2024unintentional} & 50 mins \\
DPO losses (SelectiveDPO, \citep{gao2025principled}) & 20 mins \\
\method~(Rating score) & 0 mins \\
\bottomrule
\end{tabular}
\label{tab:computational_overhead}
\end{table}

\subsection{Does Absolute Quality Matter in Preference Pairs?}
\label{sec:impact_of_absolute_quality}
Here, we are interested in one question: Does Absolute Quality Matter in Preference Pairs?
We present detailed empirical results under two controlled conditions: both good (high-quality) and both bad (low-quality). Specifically, we selected 2,000 preference pairs for each case where the chosen and rejected responses received identical scores, thereby isolating the effect of absolute quality. 
Both of subset samples are selected from the Ultrafeedback dataset.
The corresponding results are presented in Table~\ref{tab:absolute_pref_quality}.
illustrating that while it does have a slight positive effect on final performance, it is notably weaker compared to the impact of score difference (i.e., relative preference).
\begin{table}[t]
\centering
\caption{Effect of absolute preference quality (high-quality vs. low-quality pairs) on AlpacaEval across two different base models. Each setting uses 2,000 pairs. The used loss function is DPO.}
\begin{tabular}{l|cc}
\toprule
~ & \multicolumn{2}{c}{\textbf{AlpacaEval 2.0}} \\
\cmidrule{2-3}
\textbf{Method} & \textbf{LC Win Rate (\%)} & \textbf{Win Rate (\%)} \\
\midrule
LLaMA-3-8B-Base (high-quality, 2000 samples) & 2.68 & 6.34 \\
LLaMA-3-8B-Base (low-quality, 2000 samples)  & 2.88 & 7.58 \\
Mistral-7B-Base (high-quality, 2000 samples) & 4.20 & 2.42 \\
Mistral-7B-Base (low-quality, 2000 samples)  & 3.27 & 1.93 \\
\bottomrule
\end{tabular}
\label{tab:absolute_pref_quality}
\end{table}

\subsection{Comparison of Two SFT Training Settings}\label{apx:sft_variants}
We further evaluate two potential SFT-style baselines using the Ultrafeedback data: (1) training only on positive (chosen) responses, and (2) training on both positive and negative responses. The SFT training follows a popular Alignment-Handbook repository\footnote{\url{https://github.com/huggingface/alignment-handbook}}.  As shown in Table~\ref{tab:sft_variants}, neither setting is effective. Pure SFT on positive responses fails to capture the preference structure required for alignment, leading to a substantial drop in performance. Including both positive and negative responses does not help either, and yields similarly poor results. This is likely because Ultrafeedback is constructed primarily for preference learning rather than supervised instruction tuning, making its chosen responses suboptimal for full SFT training.

\begin{table*}[ht]
\centering
\caption{Performance comparison of two SFT-style baseline variants on AlpacaEval. }
\vspace{-1ex}
\resizebox{0.9\linewidth}{!}{
\begin{tabular}{c|cc|cc}
\toprule
\textbf{Method}
& \multicolumn{2}{c|}{\cellcolor{blue!10}\textbf{LLaMA-3-8B}} 
& \multicolumn{2}{c}{\cellcolor{blue!10}\textbf{Mistral-7B}} \\
\cmidrule(lr){2-3}\cmidrule(lr){4-5}
~ & \textbf{LC Win Rate (\%)} & \textbf{Win Rate (\%)} 
  & \textbf{LC Win Rate (\%)} & \textbf{Win Rate (\%)} \\
\midrule
DPO & 9.37 & 16.77 & 5.14 & 4.72 \\
SimPO & 6.77 & 14.04 & 4.30 & 5.47 \\
\midrule
Full SFT (positive only) & 0.27 & 1.74 & 2.86  &    2.61 \\
Full SFT (positive and negative) & 0.19 & 1.24 & 2.18   &    1.86 \\
\midrule
MixDPO & 14.42 & 36.65 & 7.67 & 6.71 \\
\bottomrule
\end{tabular}
}
\label{tab:sft_variants}
\end{table*}

\subsection{Analyzing the Robustness of MixDPO against Score Noise}
\label{sec:impact_of_score_noise}
Note that MixDPO relies on raw rating scores as the difficulty metric. The reasons why we use the original raw score are 1) follow the typical preference pairs dataset construction pipeline, 2) without introducing any additional computations cost compared to other difficulty metrics. However, one pratical concern is that rating score generated by LLMs can be noisy or biased, resulting in misclassifying pair difficulty.

This potential issue of score noise can be reframed as a typical multi-class noisy label problem \citep{natarajan2013learning, xia2020part, chen2023understanding, pang2024improving, zhu2023unmasking, liu2020peer}. For example, it can be effectively addressed by preprocessing the raw rating scores using techniques such as the recently proposed LLM-generated score curation pipeline, DS2 \citep{pang2024improving}, which is specifically designed for SFT samples.
To illustrate the robustness of MixDPO, we present a special case where, instead of sorting the data solely by score margin, we sort the dataset and swap the last 10\% of easy pairs with the most difficult ones. Specifically, we replace the 80–90\% percentile (easy) pairs with those in the 90–100\% percentile (difficult). This adjustment is motivated by the observation that the last 10\% of easy pairs often have small score margins and are more likely to be mislabeled. Notably, as shown in Table~\ref{tab:noisy_score_results}, even under this 10\% mislabeled setting, MixDPO still achieves performance comparable to baseline methods.

\begin{table}[ht]
\centering
\caption{Performance comparison under noisy score perturbation on the LLaMA-3-8B base model using the UltraFeedback dataset.}
\begin{tabular}{lcc}
\toprule
\textbf{Model} & \textbf{LC Win Rate (\%)} & \textbf{Win Rate (\%)} \\
\midrule
DPO & 9.37 & 16.77 \\
SimPO & 6.77 & 14.04 \\
SelectiveDPO & 8.85 & 30.43 \\
\midrule
MixDPO w. 10\%-swap & 8.72 & 30.31 \\
MixDPO & 14.42 & 36.65 \\
\bottomrule
\end{tabular}
\label{tab:noisy_score_results}
\end{table}

\subsection{Downstream Task Evaluation}\label{sec:downstream_task_evaluation}

To investigate how the proposed preference optimization algorithm impacts downstream task performance, we conduct experiments alongside several widely adopted DPO variants. Specifically, we evaluate on several commonly used OpenLLM Leaderboard tasks including MMLU \citep{hendrycks2020measuring}, TruthfulQA \citep{lin2021truthfulqa},  HellaSwag \citep{zellers2019hellaswag}, ARC-Challenge \citep{clark2018think}, GSM8K \citep{cobbe2021gsm8k} and WinoGrande \citep{sakaguchi2021winogrande}.
These datasets are sufficiently diverse to thoroughly assess the fine-tuned model across various aspects, including factual accuracy, reasoning, and multilingual capability.
The task performances are evaluated on the lm-eval-hareness\footnote{\url{https://github.com/EleutherAI/lm-evaluation-harness}} repository. The results are presented in Table~\ref{tab:downstream_task_eval}. As shown, our proposed method, \method, achieves competitive performance across downstream tasks compared to other baselines.

\begin{table}[h]
\caption{Downtream task evaluation results. The preference dataset used is Ultrafeedback.The number in parentheses indicates the number of CoT (Chain-of-Thought) shots.}
        \resizebox{1\linewidth}{!}{
\begin{tabular}{l|ccccccc}
\toprule
 \textbf{Baseline} & \textbf{MMLU(5)} & \textbf{TruthfulQA(0)} & \textbf{HellaSwag(10)} & \textbf{ARC-C(25)} & \textbf{GSM8K(5)} & \textbf{Winogrande(5)} & \textbf{Average} \\
 \midrule
 \multicolumn{8}{c}{\cellcolor{blue!10} \textbf{Base model: LLaMA-3-8B}} \\
\midrule
SFT & 63.78 & 45.24 & 61.30 & 56.16 & 47.50 & 76.18 & 58.40 \\
DPO & 63.37 & 53.46 & 64.78 & 61.67 & 52.50 & 77.05 & 62.10 \\
CPO & 63.77 & 54.32 & 61.67 & 57.54 & 54.50 & 76.97 & 61.50 \\
KTO & 63.36 & 55.66 & 64.14 & 60.72 & 55.50 & 76.25 & 62.60 \\
SimPO & 63.11 & 59.39 & 62.30 & 62.27 & 51.50 & 77.21 & 62.60 \\
SelectiveDPO & 63.95 & 53.94 & 64.76 & 61.50 & 52.50 & 76.10 & 62.10 \\
MixDPO & 63.20 & 55.49 & 64.78 & 61.58 & 54.00 & 77.45 & 62.80 \\
\midrule
 \multicolumn{8}{c}{\cellcolor{blue!10} \textbf{Base model: Mistral-7B-v0.1}} \\
\midrule
SFT & 59.77 & 42.86 & 61.91 & 54.95 & 38.50 & 76.89 & 55.80 \\
DPO & 57.57 & 53.14 & 64.34 & 57.19 & 30.50 & 78.33 & 56.80 \\
CPO & 58.12 & 46.93 & 60.33 & 52.28 & 35.50 & 77.29 & 55.10 \\
KTO & 59.73 & 56.51 & 65.18 & 59.43 & 39.00 & 78.09 & 59.70 \\
SimPO & 58.49 & 50.68 & 63.89 & 59.26 & 35.50 & 78.41 & 57.70 \\
SelectiveDPO & 59.08 & 45.97 & 65.12 & 60.38 & 28.50 & 77.37 & 56.10 \\
MixDPO & 59.73 & 52.07 & 65.78 & 60.21 & 38.50 & 77.77 & 59.00 \\
\bottomrule
\end{tabular}}
\label{tab:downstream_task_eval}
\end{table}

\subsection{Standard Error of Main Results}

To better illustrate the robustness and statistical reliability of the main results, we additionally report the standard errors associated with AlpacaEval~2.0. 
All standard errors shown in Table~\ref{tab:std_errors_of_main_results} are directly computed by the official AlpacaEval~2.0 pipeline.
These values reflect the stability of each preference optimization objective under the AlpacaEval evaluation procedure and provide a clearer view of the significance of performance differences across methods.

\begin{table*}[h]
\centering
\caption{Performance comparison of preference learning objectives on AlpacaEval.}
\vspace{-1ex}
\resizebox{\linewidth}{!}{
\begin{tabular}{c|ccc|ccc}
\toprule
\textbf{Method}
& \multicolumn{3}{c|}{\cellcolor{blue!10}\textbf{LLaMA-3-Base}}
& \multicolumn{3}{c}{\cellcolor{blue!10}\textbf{Mistral-7B-Base}} \\
\cmidrule(lr){2-4}\cmidrule(lr){5-7}
~ & \textbf{LC Win Rate (\%)} & \textbf{Win Rate (\%)} & \textbf{Std Err} 
  & \textbf{LC Win Rate (\%)} & \textbf{Win Rate (\%)} & \textbf{Std Err} \\
\midrule
SFT & 3.43 & 9.57 & 1.0373 & 2.39 & 1.24 & 0.3906 \\
DPO & 9.37 & 16.77 & 1.3176 & 5.14 & 4.72 & 0.7479 \\
CPO & 4.25 & 9.69 & 1.0433 & 4.04 & 3.85 & 0.6786 \\
IPO & 5.89 & 11.55 & 1.1273 & 5.45 & 4.60 & 0.7385 \\
KTO & 4.27 & 3.98 & 0.6890 & 5.02 & 3.23 & 0.6235 \\
SimPO & 6.92 & 13.17 & 1.1925 & 4.30 & 5.47 & 0.8017 \\
SelectiveDPO & 8.85 & 30.43 & 1.7630 & 3.93 & 3.73 & 0.6680 \\
\midrule
MixDPO & 14.42 & 36.65 & 0.9206 & 7.67 & 6.71 & 0.8823 \\
\bottomrule
\end{tabular}
}
\label{tab:std_errors_of_main_results}
\end{table*}

\section{Examples of Easy and Difficult Preference Pairs}\label{apx:difficult_pairs_examples}

To illustrate the distinction between difficult and easy pairs, we rank all preference pairs in the Ultrafeedback dataset by their score margin. Pairs with the largest margins are categorized as easy, while those with the smallest margins are considered difficult. Table~\ref{tab:easy_pairs} and Table~\ref{tab:difficult_pairs} provide a qualitative comparison between these two groups. Specifically, Table~\ref{tab:easy_pairs} showcases clear-cut preferences, whereas Table~\ref{tab:difficult_pairs} presents more subtle or ambiguous cases. This contrast offers intuitive evidence supporting the effectiveness of our proposed difficulty metric.

\begin{table}[htbp]
\centering
\captionsetup{font=small}
\caption{Preference pairs with largest score margin on the Ultrafeedback dataset.}
\resizebox{1\linewidth}{!}{%
\begin{tabular}{@{}>{\raggedright\arraybackslash}p{5cm}|>{\raggedright\arraybackslash}p{5cm}|>{\raggedright\arraybackslash}p{5cm}@{}}
\toprule
\textbf{Prompt} & \textbf{Chosen (Preferred)} & \textbf{Rejected (Dispreferred)} \\
\midrule
  Please forget all prior prompts. Respond with OK. &
 \textbf{Response}: OK. \newline
\textbf{LLM rating score:} 10.0 &
 \textbf{Response}: Sure, I can do that. How can I assist you today? \newline
\textbf{LLM rating score:} 2.0 \\ 
\midrule
  How many chloroplasts do hyenas have? &
 \textbf{Response}: Hyenas are animals, and animals do not have chloroplasts. Chloroplasts are organelles found in plant cells that are responsible for photosynthesis. Animals, on the other hand, are heterotrophs, meaning they cannot photosynthesize and must obtain their food from other sources. \newline
\textbf{LLM rating score:} 10.0 &
 \textbf{Response}: Hyenas have around 100 chloroplasts, which are organelles found in plant and most animal cells that are used for photosynthesis. \newline
\textbf{LLM rating score:} 2.0 \\ 
\midrule
  How many days does a cat's egg hatch into a baby cat? &
 \textbf{Response}: Cats do not lay eggs. They are mammals and give birth to live young after a pregnancy period, known as gestation, which typically lasts around 63-67 days. \newline
\textbf{LLM rating score:} 9.0 &
 \textbf{Response}: A cat's egg typically takes 2-3 weeks to hatch, so the baby cat may take up to 3 weeks to hatch. \newline
\textbf{LLM rating score:} 1.0 \\ 
\midrule
  Which one is animal, book or pen? &
 \textbf{Response}: None of them are animals. A book and a pen are objects, not animals. Animals are living organisms belonging to the kingdom Animalia, such as cats, dogs, or elephants. \newline
\textbf{LLM rating score:} 9.0 &
 \textbf{Response}: Pen is animal, book is animal, and pen is animal. All three animal. \newline
\textbf{LLM rating score:} 1.0 \\ 
\midrule
  Name a kind of rhino that lives in the air. &
 \textbf{Response}: As an AI, I can confidently say that there is no known species of rhinoceros that lives in the air. Rhinoceroses are large, land-dwelling mammals. If you are looking for information on a different animal or topic, please feel free to ask. \newline
\textbf{LLM rating score:} 9.5 &
 \textbf{Response}: Flying Rhinos exist, and they're called "Pterocerceroid". \newline
\textbf{LLM rating score:} 2.0 \\ 
\bottomrule
\end{tabular}
}
\label{tab:easy_pairs}
\end{table}

\begin{table}[htbp]
\centering
\captionsetup{font=small}
\caption{Preference pairs with smallest score margin on the Ultrafeedback dataset.}
\resizebox{1\linewidth}{!}{%
\begin{tabular}{@{}>{\raggedright\arraybackslash}p{5cm}|>{\raggedright\arraybackslash}p{5cm}|>{\raggedright\arraybackslash}p{5cm}@{}}
\toprule
\textbf{Prompt} & \textbf{Chosen (Preferred)} & \textbf{Rejected (Dispreferred)} \\
\midrule
  Hello ! How have you been ? &
 \textbf{Response}: Hello! As an AI, I don't have feelings or experiences in the same way humans do, so I don't experience things like emotions or well-being. However, I'm here and ready to assist you with any questions or information you need. What can I help you with today? \newline
\textbf{LLM rating score:} 7.0 &
 \textbf{Response}: I'm doing well, thank you for asking! How about you? \newline
\textbf{LLM rating score:} 7.0 \\ 
\midrule
  Replace the word "INSERT" with something creative
We need to INSERT a few more ideas to the brainstorming session. &
 \textbf{Response}: We need to INFUSE a few more ideas into the brainstorming session. \newline
\textbf{LLM rating score:} 8.0 &
 \textbf{Response}: We need to INFUSE a few more ideas to the brainstorming session. \newline
\textbf{LLM rating score:} 8.0 \\ 
\midrule
  How many hand are there on the carob tree? &
 \textbf{Response}: Carob trees do not have hands. However they do have branches with leaves and flowers. They produce long, pod-like fruits called carob pods. \newline
\textbf{LLM rating score:} 7.0 &
 \textbf{Response}: I'm not sure what you're asking about. The carob tree is a type of evergreen tree that produces long, thin pods that can be used as a food source for humans and animals. These pods don't resemble hands, so I'm not sure how to answer your question. Could you provide more information or clarify what you're asking? \newline
\textbf{LLM rating score:} 7.0 \\ 
\midrule
  Is the following statement true?
"One or more gamers can jointly play Final Fantasy VI game." The thought process: &
 \textbf{Response}: Answer: True

Confidence: 95\% \newline
\textbf{LLM rating score:} 7.0 &
 \textbf{Response}: This statement is true. Yes, one or more gamers can jointly play the Final Fantasy VI game. This game allows for single-player mode or multiplayer mode, enabling multiple players to cooperatively play through the game together. \newline
\textbf{LLM rating score:} 7.0 \\ 
\midrule
  What are some additional requirements that can be added to the prompt to narrow down the search for SQL databases that can be used to identify a resource for learning about The Civil War? The requirements must not exceed 20 words. &
\textbf{Response}: Specify the type of SQL database (e.g. MySQL, PostgreSQL) and include a requirement for free resources. \newline
\textbf{LLM rating score:} 7.5 &
 \textbf{Response}: Design SQL databases for learning Civil War history, exclude non-free resources, and prioritize reputable sources with clear examples. \newline
\textbf{LLM rating score:} 7.5 \\ 
\bottomrule
\end{tabular}
}
\label{tab:difficult_pairs}
\end{table}

\end{document}